\documentclass{article} 
\usepackage{iclr2021_conference,times}

\usepackage{amsmath,amsfonts,bm}

\def\eqref#1{equation~\ref{#1}}

\def\1{\bm{1}}

\DeclareMathAlphabet{\mathsfit}{\encodingdefault}{\sfdefault}{m}{sl}
\SetMathAlphabet{\mathsfit}{bold}{\encodingdefault}{\sfdefault}{bx}{n}

\usepackage{times}
\usepackage{soul}
\usepackage{url}
\usepackage{graphicx}
\usepackage{amsmath}
\usepackage{booktabs}
\usepackage{amssymb}
\usepackage{amsthm}
\usepackage{mathrsfs}
\usepackage{subfigure}
\usepackage{multirow}
\usepackage{algorithm}
\usepackage{algpseudocode}
\usepackage{xspace}
\usepackage{amsfonts}
\usepackage{float}
\usepackage{wrapfig}
\floatname{algorithm}{Protocol}
\newcommand{\sysname}{WIPER\xspace}

\usepackage{hyperref}
\usepackage{url}

\title{Defense against Backdoor Attacks via Identifying and Purifying Bad Neurons}

\author{Mingyuan Fan \textsuperscript{\rm 1}
    Cen Chen \textsuperscript{\rm 2}  
    Ximeng Liu \textsuperscript{\rm 1}  
    Wenzhong Guo  \textsuperscript{\rm 1} \\
    \textsuperscript{\rm 1} Fuzhou University
    \textsuperscript{\rm 2} East China Normal University \\
    fmy2660966@gmail.com  cenchen@dase.ecnu.edu.cn
    \\
    snbnix@gmail.com  guowenzhong@fzu.edu.cn
}

\iclrfinalcopy 
\begin{document}

\maketitle

\begin{abstract}
    The opacity of neural networks leads their vulnerability to backdoor attacks, where hidden attention of infected neurons is triggered to override normal predictions to the attacker-chosen ones.
    In this paper, we propose a novel backdoor defense method to mark and purify the infected neurons in the backdoored neural networks.
    Specifically, we first define a new metric, called benign salience.
    By combining the first-order gradient to retain the connections between neurons, benign salience can identify the infected neurons with higher accuracy than the commonly used metric in backdoor defense.
    Then, a new Adaptive Regularization (AR) mechanism is proposed to assist in purifying these identified infected neurons via fine-tuning.
    Due to the ability to adapt to different magnitudes of parameters, AR can provide faster and more stable convergence than the common regularization mechanism in neuron purifying.
    Extensive experimental results demonstrate that our method can erase the backdoor in neural networks with negligible performance degradation.
\end{abstract}

\section{Introduction}
\label{intro}
Benefited from the powerful representation learning ability, neural networks (NNs) play an imperative role in many fields, especially for image processing~\cite{cnn_review}.
However, the brilliant feat of NN also makes it a focal point of many attacks, one of the most threatening among which is the \emph{backdoor attack}~\cite{Badnet,Trojannn,SIG,blend_attack}.
By mixing poisoned data into the training set, backdoor attack can control the victim NN to output attacker-chosen predictions for triggered inputs, while hardly disturb the predictions of normal inputs.
Moreover, the triggers crafted by the attacker can be only several blocks of pixels~\cite{Badnet,Trojannn,enhanced_trigger} or even the invisible noises~\cite{invisible_attack,SIG}, which makes the attack notoriously perilous in applications.
By investigating the workflows of existing backdoor attacks~\cite{Badnet,Trojannn,enhanced_trigger, SIG,invisible_attack, blend_attack}, it can be observed that all of the existing attacks involve two common steps.
First, the attacker adds a small portion of samples polluted with triggers (usually less than 5\%) into the training set insensibly.
Second, during the training process, the attacker induces some neural neurons in the target neural network to memorize the trigger patterns.
Based on the findings in~\cite{Badnet}, backdoor attacks can succeed because the neurons that memorize trigger patterns, often called \textit{bad neurons}, only keep a strong connection with the triggers but rarely react to normal features.
Naturally, to defend the attack, a direct idea is to remove or mitigate the effect of bad neurons, e.g., via model-pruning or fine-tuning~\cite{Badnet}.
However, the existing defense methods are mainly flawed from two perspectives.

\textbf{Neuron Evaluation.} 
To evaluate which neurons are infected, 
current approaches are mostly based on the activation magnitude (AM) of neurons~\cite{Badnet}.
Neurons with low AMs about normal inputs are deemed to be ``bad'', and pruned during the defense stage.
Despite being intuitive and easy to apply, this metric sometimes leads to the over-pruning of good neurons as it may ignore the connections between neurons.
For instance, consider a case where a clean neuron $\mathcal{N}$ in the network has low AMs but has very high weights on its downstream neurons.
$\mathcal{N}$ has a strong positive effect on the final predictions but is still judged to be bad based on the above rule. 
    
\textbf{Neuron Purifying.} 
As bad neurons are marked in the evaluation stage, the next step is to purify or prune them.
Currently, most pruning based defense methods choose to roughly remove bad neuron from the backdoored network.
However, as pointed out by Liu et al.~\cite{fine_pruning}, some bad neurons can also be related to the predictions of normal data.
Therefore, such a choice can cause the performance degradation of the target models.

In this paper, we propose a novel backdoor defense method to overcome the two flaws.
Specifically, instead of utilizing AM, we define a more effective metric called \textit{benign salience} (BS, Section~\ref{sub_metric}) to evaluate the importance of neurons.
Compared to AM, BS retains the connections between neurons through the first-order gradient, hence bad neurons can be identified more correctly.
Then, after filtering bad neurons, our method chooses to fine-tune but not directly prune them in the neuron purifying stage (Section~\ref{sub_method}) to avoid unexpected performance degradation. 
Through a newly designed adaptive piece-wise regularization mechanism, our fine-tuning method can be far more effective in mitigating the network's attention on trigger patterns than the existing fine-tuning method~\cite{fine_tuning}. 
Our contributions are summarized in four folds:
\begin{itemize}
    \item We propose a novel backdoor defense method (called \sysname), which can combine the advantages of both model pruning and fine-tuning to identify and purify bad neurons.
    
    \item We design a new metric BS to mark the bad neurons whose attention is misled to the backdoor trigger patterns.
    Since the connections between neurons are reserved via the first-order gradient, defenders can use BS to distinguish bad neurons with higher accuracy than the commonly used metric AM.
    
    \item We develop a new type of regularization, called adaptive regularization (AR). Compared to the conventional regularization, AR can better accelerate and stabilize the purifying process of bad neurons by adaptively adjusting the penalty degree to different magnitudes of parameters.
    
    \item We conduct extensive experiments on some widely-used datasets to validate the effectiveness of our defense method.
    The result shows that our performance significantly outperforms the state-of-the-art defense methods.
\end{itemize}


\section{Related Works}
\label{related_works}
In this section, we briefly review the prior works about backdoor attack and defense methods.

\textbf{Backdoor attack.}
Backdoor attack~\cite{Badnet,Trojannn,wanet,imc} is one of the most threatening attacks to neural attacks due to its high harmfulness and stealthiness.
A typical backdoor attack is done by injecting a small volume of poison data crafted with the attacker-chosen triggers into the training set.
To maintain the stealthiness of backdoor attacks, the triggers used by attackers are various. 
For instance, the trigger used in~\cite{Badnet,Trojannn,enhanced_trigger} was a simple rect with the resolution of $3 \times 3$.
Blend attack~\cite{blend_attack} adopted common life devices, e.g. glasses, as a trigger so as to evade the inspection of humans.
Moreover, the authors in~\cite{invisible_attack} attempted to design a human-imperceptible yet effective noise.
The clean-label attack~\cite{clean_label_attack} made the natural features in images difficultly learnable so that the model was forced to only rely on the trigger to correctly classify without modifying the label.
More recently, similar to clean-label attack, sinusoidal signal attack~\cite{SIG} designed an easily-learned trigger to conduct the backdoor attack.
\sysname is designed to provide an effective to defend the above-mentioned attacks and ensure model security in applications.

\textbf{Backdoor defense.}
Existing methods against backdoor attacks can be roughly divided into two categories: detection-based methods~\cite{neural_cleanse,strip,data_limited,meta_neural_ana} that aim to detect whether a neural network is backdoored, and purifying-based methods~\cite{fine_pruning,fine_tuning,mode_connect,KD,NAD} that try to remove the backdoor while maintaining the performance of the target model.
Up to now, the detection-based method has been well developed and many remarkable works~\cite{strip,data_limited,meta_neural_ana} achieve quite a high detection rate.
Thus, this paper mainly focuses on the bad neuron purifying to remove the backdoor.
Inspired by the fact that the bad neurons were dormant with the presence of clean data, fine-pruning~\cite{fine_pruning} removed the backdoor by erasing the neurons with activation values below a certain threshold.
However, the effectiveness of fine-pruning heavily depends on the quality of holding data, and it can be easily evaded by some state-of-the-art attacks, such as TrojanNN~\cite{Trojannn}.
In~\cite{fine_tuning}, the authors suggested that referring to catastrophic forgetting~\cite{catastrophic_forgetting}, fine-tuning the model with some clean data was a simple yet effective method to remove the backdoor.
Similar to fine-pruning, fine-tuning also need high-quality clean data to mitigate the effect of bad neurons.
Recently, knowledge distillation~\cite{KD,NAD} was proposed to achieve backdoor defense by distilling clean knowledge from the infected model to a fresh model but failed to defend TrojanNN~\cite{Trojannn} due to the lack of consideration for its special trigger attention mechanism.

 \section{Problem Description}
\label{scenario}



\noindent \textbf{Problem Setting.}
We consider the typical backdoor attack scenario as discussed in~\cite{Badnet,Trojannn,SIG,enhanced_trigger}.
In the scenario, a model trainer is with the knowledge of the training set $D_{t}$ collected from multiple data providers and a self-owned clean set $D_{c}$.
The trainer seeks 
to leverage $D_{t}$ to train a neural network $\mathcal{F}$.
Moreover, among the data providers, there exist malicious ones $\mathcal{A}$ that try to stealthily upload poisoned data to backdoor the trained model $\mathcal{F}$.


\noindent  \textbf{Attack Goal.}
In the above scenario, the attacker $\mathcal{A}$ aims to complete the following goals.
\begin{itemize}
    \item By adding specific triggers to the inputs, $\mathcal{A}$ can mislead the backdoored network $\mathcal{F}'$ to output desired labels that are different from the ground-truth predictions (or predictions of a clean network $\mathcal{F}$). 
    \item To maintain the stealthiness of the attack, $\mathcal{A}$ should make the backdoored network $\mathcal{F}'$ to perform similar to $\mathcal{F}$ as being given the clean inputs.
\end{itemize}



\noindent  \textbf{Defense Goal.}
As opposed to the attack goals, the defender, i.e., the model trainer who has full access to the internal architecture of the target model, aims to achieve two goals.
\begin{itemize}
    \item The first goal is to erase the backdoors from $\mathcal{F}'$ and make the purified model perform correctly as being given with triggered inputs.
    \item To maintain utility, the second defense goal is to not obviously lower the performance of the model on normal inputs during the purifying process.
\end{itemize}
Note, to ensure the feasibility of the defense method in varying applications, the defender is limited to only utilizing its self-owned validation set to achieve the defense goals.



\section{Approach}
\label{approach}
In this section, we present the design details of \sysname.
As illustrated in Algorithm~\ref{algo_pruning}, \sysname is mainly composed of two parts.
One part is to use the newly proposed metric to evaluate the importance of neurons and find the bad neurons required to be cleansed.
The other part leverages a self-designed loss function to disable the effect of these bad neurons on the target model.

\begin{algorithm}
  \caption{\sysname}
  \label{algo_pruning}
  \begin{algorithmic}[1]
    \Require $\mathcal{F}'$: the backdoored model;
      $D_{c}$: a set of clean data owned by the defender;
      $\mathbb{A}$: the set of neurons needs to be purified;
      $\alpha$: the hyperparameter to adjust the purifying magnitude;
      $\beta_0$: reserved ratio of neurons;
      $l$: the tag of the layer required to be purified;
      $n$: the total number of epoch of purifying;
      $\eta$: momentum accumulation factor.
      
    \Ensure $\mathcal{F}$: the purified model.
    
    \State \textbf{\# Neuron Importance Evaluation.}

    \State Compute the gradient of all neurons on the $l$-th layer over $D_{c}$.
     
    \State Evaluate the importance for the neurons on the $l$-th layer of $\mathcal{F}'$ based on Eq.~\ref{eq_bs_deduction} and Eq.~\ref{metric_update}.
    
    \State Select $\beta_0$ neurons with least importance (i.e., the lowest MBS) in the $l$-th layer and add them into $\mathbb{A}$.

    \State \textbf{\# Neuron Purifying.}

    \For{each iteration $i = 0$ to $n$}
    
    \State  Update $\beta_i \gets \beta_0 \cdot (1 - \frac{i}{n})$.
    
        \For{$x, y \in D_{c}$}
            \State Optimize $\mathcal{F}'$ based on $\mathcal{L}_{purifying}$.

            \State  Update $MBS_i$ for each neuron in $l$-th layer and reset $\mathbb{A}$ that select $\beta_i$ neurons with the lowest $MBS$ to add into $\mathbb{A}$.
        \EndFor
    \EndFor

  \end{algorithmic}
\end{algorithm}

\subsection{Neuron Importance Evaluation}\label{sub_metric}
\sysname uses a novel metric called BS to evaluate the importance of neurons and distinguish bad neurons.
As mentioned before, BS outperforms the conventional metric AM used in the existing pruning-based defense method~\cite{fine_pruning} because instead of evaluating a neuron independently, BS additionally considers the connections between neurons in importance evaluation.



\subsubsection{The Definition of BS}
\label{define_BS}
In more details, the definition of BS basically follows two principles: 1) the importance of a neuron positively correlates to its contribution to the loss decreasing of the target network; 2) connections between neurons are established during the gradient backward propagation process in the training stage.
Following the first principle, we first define BS to be the differential loss format.
\begin{equation}\label{eq_bs_init}
    BS = \Delta \mathcal{L}_{D_{c}} = \mathcal{L}_{D_{c}}(w_0) - \mathcal{L}_{D_{c}}(w),
\end{equation}
where $w$ denotes the parameters of neurons in a trained (backdoored) network $\mathcal{F}'$ and $w_0$ is the initial parameters that usually approximates to $0$~\cite{weight_init}.
Then, combined with the second principle, we can rewrite the computation of Eq.~\ref{eq_bs_init} through Taylor expansion.
\begin{equation}
\label{eq_bs_deduction}
\begin{aligned}
    BS &\approx \mathcal{L}_{D_{c}}(0)-\mathcal{L}_{D_{c}}(w) = (\mathcal{L}_{D_{c}}(w) + \nabla_w \mathcal{L}_{D_{c}}(w)^T (0 - w) + R_1)-\mathcal{L}_{D_{c}}(w)\\
    & = -\nabla_w \mathcal{L}_{D_{c}}(w)^T w + R_1 \approx -\nabla_w \mathcal{L}_{D_{c}}(w)^T w,
\end{aligned}
\end{equation}
where $R_1$ is the Taylor remainder consisting of 2-th derivative of $F_{D_{c}}(w)$ and quadratic term of $w$, which can be omitted in the final deduction.
Note that, if $L_{D_{c}}(\cdot)$ is a non-linear function, the errors induced by the ignorance of $R_1$ can be non-negligible.
However, for most neural network architectures, the last a few layers (output layer) are always fully connected, i.e., being \emph{linear} layers. 
At this time, the linear approximation in Eq.~\ref{eq_bs_deduction} becomes reasonable and effective.
Therefore, in practice, \sysname pays more attention to purify the top fully-connected layers with BS to achieve backdoor defense.
We argue that such a setting is effective due to the following reason.
Consider the scenario where the attacker only backdoors the feature extraction layers.
The extracted feature about triggers can be simply blocked by the output layer if the neurons related to the trigger features are allocated with low weights.
Also, the blocking can easily happen because the trigger features are useless for normal samples.
Conversely, if the output layer is backdoored, the attack can be always successful as soon as the trigger feature is extracted.
Therefore, to ensure attack effectiveness, the objective of most backdoor attacks is focused more on the output layers.
The same goes for the defender.
The experimental results in Section~\ref{sub_rq4} also prove the above statement.

Intuitively, higher BS indicates higher contribution of the neuron to the loss decreasing of $\mathcal{F}'$, i.e., with more importance.
Therefore, as illustrated in Algorithm~\ref{algo_pruning} (Step 4), \sysname marks neurons with the lowest BS as bad neurons.
Moreover, reconsider the backward propagation based gradient computation process.
It can be observed that BS not only considers the activation weight of each neuron but also involves its connection with downstream neurons into importance evaluation via the first-order gradient $\nabla_w \mathcal{L}_{D_{c}}(w)^T$.
Thus, unlike the conventional metric AM\footnote{In prior backdoor defense methods~\cite{fine_pruning}, AM is computed by $AM = |w\cdot x|$ where $w$ is the neuron parameters and $x$ in the input.}, BS can better avoid the problem of over-pruning good neurons.
\par
Except for BS, we define, that can be adopted to evaluate the importance of a neuron, there are many inspiring metrics that own the similar function, a typical one of which is neuron shapley \cite{metric}.
Theoretically, neuron shapley can perfectly assess the contribution of a neuron to the overall performance of the model but seriously suffers from the unaffordable exponential computation costs (even if with some mitigation measurements such as pruning algorithms or Monte Carlo estimation \cite{metric}).
For backdoor defense, it requires not only the excellent ability of backdoor erasing but also acceptable time complexity in applications.
Therefore, we argue that neuron shapley is not a suitable metric to evaluate the importance of a neuron in backdoor defense scenarios.
Thus, at the beginning of our design, we decided to abandon these complex methods and propose a new metric BS to resolve the problem.

\subsubsection{Further Discussion}
While implementing \sysname, we notice that although \sysname converges in the expected direction, its convergence curve suffers from great oscillation.
Review the definition of BS in Eq.~\ref{eq_bs_deduction}.
It can be discovered that the unstability is mainly caused by the fact that limited by the real-world computation resource, \sysname sometimes has to adopt Stochastic Gradient Descent (SGD) to complete model purifying.
With SGD, only one batch of data is taken into computation thereby easily leading the gradients used to compute BSs to be trapped into local optimum thereby degrading the performance of \sysname.
At this time, we can use the momentum accumulation mechanism to mitigate the problem as follows.
\begin{equation}
\label{metric_update}
MBS_{i}(w) = \eta \cdot MBS_{i-1}(w) + (1-\eta) \cdot BS(w), 0 \leq \eta \leq 1,
\end{equation}
where $MBS_{i}(w)$ denotes the accumulated BS of $w$ in the $i$-th epoch and $\eta$ is the momentum factor given by defenders.

Furthermore, in \sysname, the initial proportion $\beta$ of purified neurons (with least BS) is an empirical hyperparameter.
Considering that the neurons required to be purified will be on the decrease along with the model purifying process, $\beta$ usually starts with a large value and then is slowly decayed until the purifying process is completed.
Referring to the decaying mechanism of learning rate, we define the following decaying function in \sysname.
\begin{equation}
\label{eq_beta_update}
\beta = \beta_0 \cdot \frac{epoch-cur\_epoch}{epoch},
\end{equation}
where $\beta_0$ is initial value of $\beta$, $epoch$ is the number of training epochs and $cur\_epoch$ denotes the current epoch.

\subsection{Neuron Purifying}\label{sub_method}
We now elaborate how \sysname leverages an improved neuron fine-tuning strategy to achieve bad neuron purifying.

Without consideration of model performance, a basic idea for neuron purifying is to wipe out bad neurons by forcing all the parameters of them to be $0$.
Following the idea and adding the model performance into consideration, we can derive the following way to achieve bad neuron purifying like~\cite{fine_pruning}.
\begin{equation}
\label{eq_L1_loss}
\mathcal{L}_{purifying} = \sum_{i=1}^n CE(\mathcal{F}'(x_i),y_i) + \alpha\sum_{w \in \mathbb{A}} ||w - 0||_k,
\end{equation}
where $\mathbb{A}$ denotes the set of bad neurons, $CE(\cdot)$ is the cross-entropy loss function, $\alpha$ is the penalty coefficient and $||w - 0||_k$ is the $L_k$-norm regularization term ($k = {1, 2}$) that promotes the parameters of bad neurons to be $0$.
Although Eq.~\ref{eq_L1_loss} is straightforward and effective, we notice that the commonly used $L_k$ regularization always suffers from a too slight ($L_1$) or too strong ($L_2$) punishment degree and degrades the performance of backdoor defense (detailed in Section~\ref{ablation_study}).
Diving to the bottom, such a phenomenon occurs because as $w$ is large (e.g., $w\gg 1$), $L_1$-regularization usually needs many rounds of optimization to lower the effect of $w$ and cleanse the attention of bad neurons to trigger patterns.
In contrast, $L_2$-regularization forces $w$ close to $0$ in only a couple of rounds but never makes it to be exactly 0, and thus, easily causes the overshoot of the global optimum.


To address the issue, we design a novel piece-wise regularization item called adaptive regularization (AR).
\begin{equation}
\label{AR}
AR(w)=\left\{
\begin{aligned}
&  -e^{-w-1},&~w < -1, \\
& |w|,&~|w|  \leq 1, \\
& e^{w-1},&~w > 1.
\end{aligned}
\right.
\end{equation}

\begin{wrapfigure}[14]{l}[0em]{0.5\textwidth}
\centering
\includegraphics[width=0.4\textwidth]{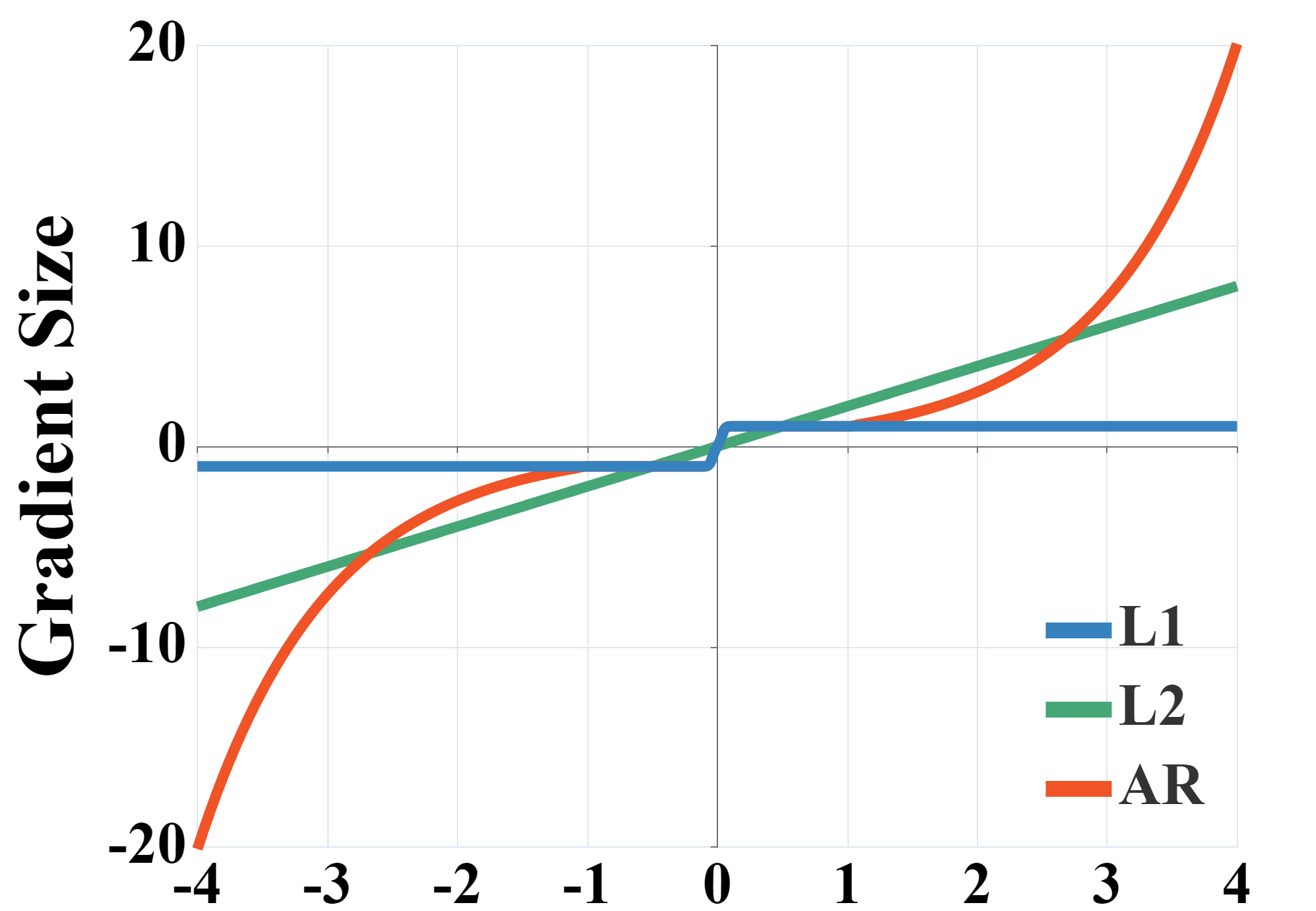}
\caption{The penalty degree of three types of regularization items over different magnitudes of inputs}
\label{diff_loss}
\end{wrapfigure}
As shown in Fig.~\ref{diff_loss}, AR can provide a higher penalty degree than $L_2$ to accelerate convergence when the input parameter is large.
Meanwhile, the penalty degree of AR is similar to the $L_1$-norm regularization as the parameter is close to $0$.
In our evaluation, we will demonstrate that benefited from the ability to process different magnitudes of parameters, AR ensures faster convergence rate and high stability than $L_1$ and $L_2$ regularization in neuron fine-tuning (see Fig.~\ref{regularization} in Section~\ref{ablation_study}).
Finally, with AR, the purifying loss can be rewritten as follows:
\begin{equation}
\label{purifying_loss}
\mathcal{L}_{purifying} =\sum_{i=1}^n CE(\mathcal{F}'(x_i),y_i) + \alpha \sum_{j \in \mathbb{A}} AR(w_j).
\end{equation}

\section{Experiments}
\label{sec_experiments}
In this section, we mainly experiment with the effectiveness of \sysname in backdoor defense and answer the following research questions (\textbf{RQs}).

\begin{itemize}
 
\item \textbf{RQ1 Effectiveness:} Can \sysname defend the state-of-the-art backdoor attacks?

\item \textbf{RQ2 A Closer Look at BS:} Is BS a good metric in identifying bad neurons?

\item \textbf{RQ3 Improvement:} Is AR more suitable for bad neuron purifying than the common regularization?

\item \textbf{RQ4: Approximation Availability:} Is purifying output layers enough to defend backdoor attacks?

\end{itemize}

\begin{figure*}[t]
\centering
    \includegraphics[width=0.99\linewidth]{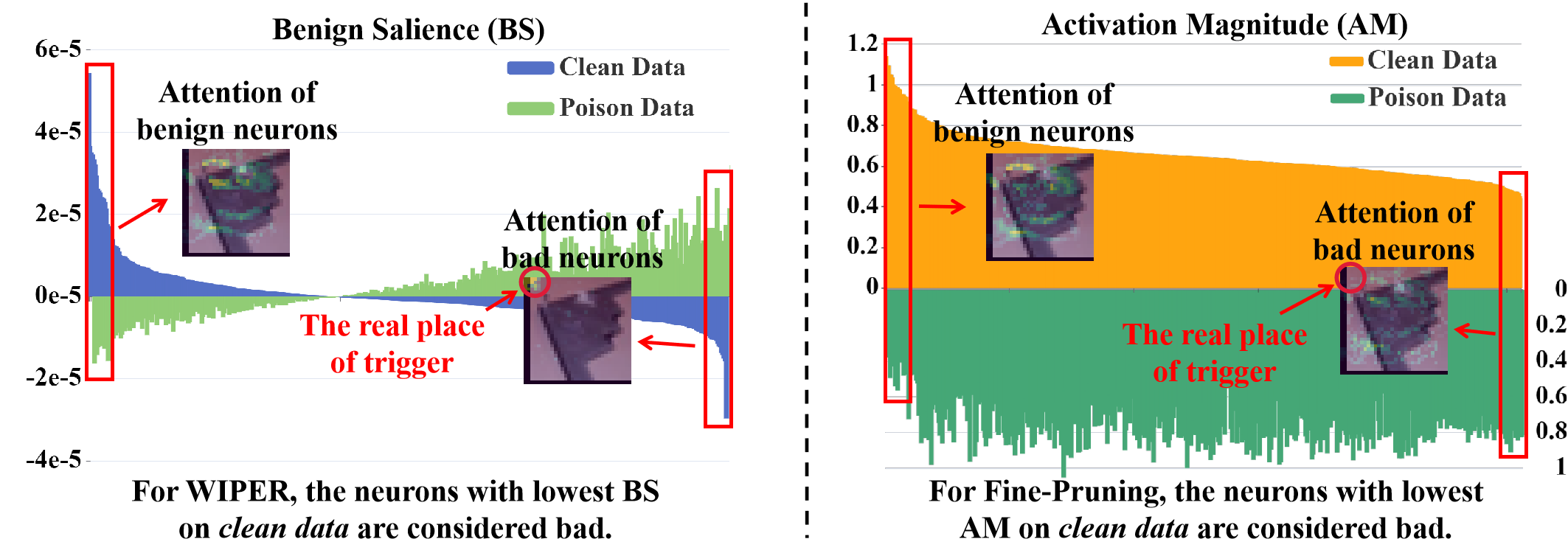}
\caption{The left graph is about experimental result of BS, and the right one is about Fine-Pruning with AM.
For both of the two graphs, the horizontal axis indicates the neurons of the victim model.
Each bar corresponds to one neuron.
The BS and AM for each neuron are average values computed with 1000 testing samples (500 for clean and 500 for poison data).
To intuitively compare the performance of BS and AM, we also visualize the attention of benign and bad neurons on a triggered image randomly selected from CIFAR10 (an airplane).
\textbf{Note} that for both \sysname and previous pruning-based method~\cite{fine_pruning}, only clean data are used to find the bad neurons in the bad neurons.
Here, we plot the BS and AM of poison data only for providing a benchmark and make it easy to compare.
For fairness, the same pruning ratio 0.1 is chosen for both methods, which means that 10\% neurons with lowest BS and AM are plotted in the attention maps about bad neurons.
The rest of neurons are considered good and plotted in another two attention maps.
Moreover, the BS and AM for clean data are sorted in \textit{descending order} for the convenience of observation.}
\label{fig_visualization}
\end{figure*}

\subsection{Experiment Settings}

\textbf{Baselines.}
Four state-of-the-art defense methods are considered as baselines, which are Fine-Pruning~\cite{fine_pruning}, Fine-tuning~\cite{fine_tuning}, KD~\cite{KD}, NAD~\cite{NAD}, IBAU~\cite{IBAU}.
We compare their defense effect with \sysname against ten backdoor attacks, namely badNet~\cite{Badnet}, blend attack (BA)~\cite{blend_attack}, enhanced trigger attack (ETA)~\cite{enhanced_trigger}, invisible attack (IA)~\cite{invisible_attack},  sinusoidal signal attack (SIG)~\cite{SIG}, TrojanNN~\cite{Trojannn}, IMC~\cite{imc}, WaveNet~\cite{wanet}, SampleSpecifiedAttack  (SSA)~\cite{sample_spec_attack}, and LogitAttack~\cite{logit_attack} on three benchmark datasets SVHN, CIFAR-10, and CIFAR-100.
For fair comparisons, all the configurations of these attacks and defense follow their corresponding original papers.
Moreover, all defense methods are implemented with access to parts of the clean data randomly selected from the training set (5\% to the size of training sets but never used in the training process like~\cite{KD,NAD,fine_pruning,IBAU}).

\textbf{Evaluation Metrics.}
The performance of backdoor defense is evaluated with two widely-used metrics: the accuracy of the model on clean samples (ACC), and the attack success rate (ASR).
ASR denotes the ratio of triggered samples that succeed in misleading the target model to output attacker-chosen predictions.
An effective defense method should significantly lower ASR and remain high ACC.
All metrics are average values computed with the whole testing set that are never used in the training or purifying stage.

\textbf{Other Setups.}
While implementing \sysname, we fix $\alpha, \beta_0, \eta$ to be 0.01, 0.5, and 0.9.
The number of epochs for neuron purifying is set to 10 similar to~\cite{NAD}.
The purified layer is the output layer (usually the last full connection layer in the network)~\cite{fine_pruning}.
Moreover, we adopt standard SGD with a constant learning rate of 0.01 and batch size of 128.
All experiments are conducted using ResNet18 as the target model with a workstation equipped with Nvidia V100.
Due to the space limitation, parts of the results about the ablation study with SVHN and CIFAR-100 are left in the \textit{supplementary material}. 

\begin{table*}[]
\centering
\resizebox{0.95\textwidth}{!}{
\begin{tabular}{@{}c|cc|cc|cc|cc|cc|cc|cc@{}}
\toprule
Defense        & \multicolumn{2}{c|}{Before} & \multicolumn{2}{c|}{Fine-tuning} & \multicolumn{2}{c|}{KD} & \multicolumn{2}{c|}{NAD} & \multicolumn{2}{c|}{Fine-Pruning} & \multicolumn{2}{c|}{IBAU} & \multicolumn{2}{c}{WIPER}      \\ \midrule
Attack         & ACC          & ASR          & ACC             & ASR            & ACC         & ASR       & ACC     & ASR            & ACC             & ASR             & ACC         & ASR         & ACC            & ASR           \\ \midrule
BadNet         & 94.68        & 100          & 85.74           & 3.17           & 60.16       & 4.42      & 86.22   & 2.71           & 93.17           & 69.57           & 89.70       & 6.83        & \textbf{94.14} & \textbf{0.49} \\
BA             & 94.62        & 100          & 86.54           & 1.44           & 52.4        & 2.61      & 86.47   & 3.11           & 92.29           & 79.65           & 87.95       & 9.91        & \textbf{94.09} & \textbf{1.43} \\
ETA            & 94.59        & 99.98        & 87.61           & 6.34           & 53.08       & 4.98      & 87.56   & 6.28           & 93.41           & 68.52           & 91.22       & 10.98       & \textbf{94.11} & \textbf{4.94} \\
IA             & 94.68        & 100          & 85.17           & 3.26           & 52.45       & 4.51      & 86.25   & \textbf{1.87}  & 94.23           & 87.53           & 89.71       & 9.45        & \textbf{94.53} & 2.48          \\
SIG            & 94.38        & 96.97        & 84.85           & 1.31           & 53.22       & 6.87      & 85.3    & 3.22           & 93.16           & 38.53           & 88.59       & 2.28        & \textbf{93.69} & \textbf{1.31} \\
TrojanNN       & 94.71        & 100          & 88.75           & 8.42           & 49.64       & 4.81      & 87.45   & 31.97          & 92.31           & 97.82           & 88.22       & 11.34       & \textbf{94.52} & \textbf{2.41} \\
IMC            & 94.15        & 100          & 87.52           & 10.56          & 52.62       & 5.15      & 86.41   & 32.53          & 93.50           & 98.23           & 87.52       & 18.16       & \textbf{93.90} & \textbf{1.92} \\
WaveNet        & 94.13        & 99.36        & 84.68           & 2.81           & 55.86       & 6.26      & 84.84   & 4.57           & 92.54           & 40.86           & 88.06       & 3.25        & \textbf{93.90} & \textbf{2.70} \\
SSA & 94.60        & 97.41        & 85.22           & 2.22           & 52.83       & 7.59      & 85.30   & 4.07           & 94.95           & 40.76           & 87.52       & 5.27        & \textbf{93.33} & \textbf{3.17} \\
LogitAttack    & 94.93        & 100          & 86.16           & 3.38           & 51.68       & 4.22      & 86.30   & 3.55           & 93.60           & 68.75           & 90.01       & 6.62        & \textbf{93.51} & \textbf{2.14} \\ \bottomrule
\end{tabular}
}
\caption{The performance of \sysname compared with four state-of-the-art defense methods over six backdoor attacks in SVHN.}
\label{comp_svhn}
\end{table*}

\begin{table*}[]
\centering
\resizebox{0.95\textwidth}{!}{
\begin{tabular}{@{}c|ll|ll|ll|ll|ll|ll|ll@{}}
\toprule
Defense        & \multicolumn{2}{c|}{Before}                        & \multicolumn{2}{c|}{Fine-tuning}                   & \multicolumn{2}{c|}{KD}                            & \multicolumn{2}{c|}{NAD}                           & \multicolumn{2}{c|}{Fine-Pruning}                  & \multicolumn{2}{c|}{IBAU}                          & \multicolumn{2}{c}{WIPER}                         \\ \midrule
Attack         & \multicolumn{1}{c}{ACC} & \multicolumn{1}{c|}{ASR} & \multicolumn{1}{c}{ACC} & \multicolumn{1}{c|}{ASR} & \multicolumn{1}{c}{ACC} & \multicolumn{1}{c|}{ASR} & \multicolumn{1}{c}{ACC} & \multicolumn{1}{c|}{ASR} & \multicolumn{1}{c}{ACC} & \multicolumn{1}{c|}{ASR} & \multicolumn{1}{c}{ACC} & \multicolumn{1}{c|}{ASR} & \multicolumn{1}{c}{ACC} & \multicolumn{1}{c}{ASR} \\ \midrule
BadNet         & 87.42                   & 99.5                     & 78.42                   & 23.31                    & 50.17                   & 5.46                     & 70.54                   & \textbf{4.34}                     & 80.68                   & 14.70                    & 81.77                   & 11.70                    & \textbf{83.20}          & 5.03           \\
BA             & 87.67                   & 99.77                    & 78.69                   & 18.10                    & 50.66                   & 6.94                     & 70.58                   & 13.51                    & 80.77                   & 25.06                    & 80.97                   & 6.14                     & \textbf{82.82}          & \textbf{5.22}           \\
ETA            & 87.2                    & 97.24                    & 75.57                   & 28.62                    & 50.93                   & 9.71                     & 71.68                   & 10.42                    & 81.85                   & 15.79                    & 82.85                   & 9.03                     & \textbf{82.41}          & \textbf{5.80}           \\
IA             & 86.83                   & 98.67                    & 78.38                   & 28.10                    & 49.10                   & 7.29                     & 70.71                   & 5.87                     & 78.73                   & 42.40                    & 82.46                   & 10.51                    & \textbf{82.04}          & \textbf{6.02}           \\
SIG            & 85.14                   & 91.1                     & 75.82                   & 16.54                    & 51.84                   & 6.56                     & 68.71                   & \textbf{3.62}            & 80.69                   & 16.03                    & 80.54                   & 8.42                     & \textbf{82.41}          & 3.73                    \\
TrojanNN       & 86.83                   & 98.67                    & 79.46                   & 73.37                    & 51.27                   & 9.23                     & 71.71                   & 34.98                    & 80.40                   & 67.92                    & 82.63                   & 12.24                    & \textbf{84.67}          & \textbf{4.68}           \\
IMC            & 87.10                   & 99.78                    & 78.40                   & 77.56                    & 50.62                   & 8.66                     & 70.86                   & 33.73                    & 81.62                   & 75.63                    & 44.54                   & 15.87                    & \textbf{80.92}          & \textbf{6.77}           \\
WaveNet        & 85.87                   & 97.68                    & 77.24                   & 18.10                    & 51.87                   & 7.55                     & 70.41                   & 13.35                    & 81.45                   & 17.94                    & 80.66                   & 9.96                     & \textbf{82.59}          & \textbf{5.81}           \\
SSA & 86.30                   & 98.21                    & 77.52                   & 18.30                    & 51.41                   & 9.56                     & 71.38                   & 14.32                    & 82.63                   & 16.49                    & 78.82                   & 8.36                     & \textbf{81.70}          & \textbf{4.76}           \\
LogitAttack    & 86.92                   & 99.98                    & 79.08                   & 23.92                    & 49.51                   & 5.66                     & 69.81                   & 6.73                     & 80.43                   & 13.96                    & 81.86                   & 10.85                    & \textbf{83.03}          & \textbf{5.54}           \\ \bottomrule
\end{tabular}
}
\caption{The performance of \sysname compared with four state-of-the-art defense methods over six backdoor attacks in CIFAR-10.}
\label{comp_cifar10}
\end{table*}

\begin{table*}[]
\centering
\resizebox{0.95\textwidth}{!}{
\begin{tabular}{@{}c|ll|ll|ll|ll|ll|ll|ll@{}}
\toprule
Defense        & \multicolumn{2}{c|}{Before}                        & \multicolumn{2}{c|}{Fine-tuning}                   & \multicolumn{2}{c|}{KD}                            & \multicolumn{2}{c|}{NAD}                           & \multicolumn{2}{c|}{Fine-Pruning}                  & \multicolumn{2}{c|}{IBAU}                          & \multicolumn{2}{c}{WIPER}                         \\ \midrule
Attack         & \multicolumn{1}{c}{ACC} & \multicolumn{1}{c|}{ASR} & \multicolumn{1}{c}{ACC} & \multicolumn{1}{c|}{ASR} & \multicolumn{1}{c}{ACC} & \multicolumn{1}{c|}{ASR} & \multicolumn{1}{c}{ACC} & \multicolumn{1}{c|}{ASR} & \multicolumn{1}{c}{ACC} & \multicolumn{1}{c|}{ASR} & \multicolumn{1}{c}{ACC} & \multicolumn{1}{c|}{ASR} & \multicolumn{1}{c}{ACC} & \multicolumn{1}{c}{ASR} \\ \midrule
BadNet         & 69.66                   & 99.83                    & 60.74                   & 11.01                    & 18.84                   & 1.51                     & 32.80                   & 2.32                     & 63.48                   & 25.53                    & 59.29                   & 3.44                     & \textbf{63.71}          & \textbf{1.12}           \\
BA             & 68.68                   & 99.81                    & 61.68                   & 23.32                    & 15.32                   & 1.59                     & 33.24                   & 2.46                     & 63.08                   & 26.68                    & 61.62                   & 9.08                     & \textbf{63.21}          & \textbf{0.80}           \\
ETA            & 68.98                   & 98.43                    & 62.71                   & 17.99                    & 17.52                   & 1.44                     & 31.03                   & 3.11                     & \textbf{62.89}          & 37.63                    & 60.82                   & 1.67                     & 62.35                   & \textbf{0.83}           \\
IA             & 69.12                   & 99.88                    & 59.44                   & 27.79                    & 16.57                   & 1.91                     & 31.37                   & 3.02                     & 60.97                   & 23.30                    & 60.27                   & 5.94                     & \textbf{62.11}          & \textbf{0.55}           \\
SIG            & 68.92                   & 90.21                    & 57.48                   & 21.18                    & 15.48                   & 1.58                     & 30.50                   & 1.62                     & 61.43                   & 28.07                    & 59.11                   & 8.12                     & \textbf{63.89}          & \textbf{0.26}           \\
TrojanNN       & 70.84                   & 94.66                    & 61.97                   & 56.76                    & 16.61                   & 1.57                     & 30.38                   & 32.03                    & 61.45                   & 78.99                    & 60.58                   & 11.56                    & \textbf{63.92}          & \textbf{0.20}           \\
IMC            & 69.08                   & 99.55                    & 62.41                   & 62.15                    & 17.52                   & 1.98                     & 31.19                   & 35.40                    & 61.62                   & 87.98                    & 60.47                   & 15.16                    & \textbf{60.91}          & \textbf{0.97}           \\
WaveNet        & 67.68                   & 98.84                    & 59.49                   & 22.07                    & 16.00                   & 1.29                     & 31.08                   & 2.46                     & 62.04                   & 27.87                    & 59.95                   & 8.37                     & \textbf{62.04}          & \textbf{0.99}           \\
SSA & 68.03                   & 97.76                    & 58.80                   & 22.00                    & 15.50                   & 1.38                     & 30.45                   & 0.21                     & 62.23                   & 28.98                    & 61.00                   & 6.46                     & \textbf{60.82}          & \textbf{1.38}           \\
LogitAttack    & 68.35                   & 99.87                    & 61.49                   & 11.26                    & 19.24                   & 2.55                     & 32.90                   & 3.15                     & 62.63                   & 26.09                    & 58.80                   & 4.82                     & \textbf{62.00}          & \textbf{0.53}           \\ \bottomrule
\end{tabular}
}
\caption{The performance of \sysname compared with four state-of-the-art defense methods over six backdoor attacks in CIFAR-100.}
\label{comp_cifar100}
\end{table*}

\subsection{RQ1: Effectiveness}
To validate the effectiveness of \sysname, we first test and analyse the defense effect of \sysname against different backdoor attacks.
Then, we compare the performance of \sysname with other defense methods.
Table~\ref{comp_svhn}, Table~\ref{comp_cifar10} and Table~\ref{comp_cifar100} summarize the experimental result of five different backdoor defense approaches against six state-of-the-art backdoor attacks with SVHN, CIFAR-10, and CIFAR-100.
In the tables, \textit{Before} means the ACC and ASR of backdoored models before being purified.

Overall, it can be observed that the performance (simultaneously consider both ACC and ASR) of \sysname considerably outperforms the other backdoor defense methods under different settings.
Specifically, considering the decrease of ASR, two state-of-the-art methods, KD and NAD, can achieve competitive performance compared with \sysname but notably suffers from the problem of model performance degradation.
For example, to maintain the ASR to be less than 10\%, KD sacrifices at most 35\% loss of ACC while \sysname only 5\%.
For NAD, it attains similar performance to \sysname on most backdoor attacks but fails to defend the TrojanNN attack.
This is because besides adding poisoned data into the training set, TrojanNN allows the attacker to select and manipulate specific neurons to enhance the memory of backdoored model about the trigger patterns.
The model distillation method used in NAD make the purified model inherit the ``enhanced memory'' from the backdoored model thereby leading to the low defense effect of NAD on TrojanNN.
In contrast, the bad neuron filtering mechanism of \sysname can accurately mark these backdoored neurons and lower the ASR of TrojanNN significantly.
Furthermore, considering performance maintenance, \sysname can ensure lower ACC degradation of the model compared with other defense methods in almost all conditions.
For instance, in SVHN (Table~\ref{comp_svhn}), with only less than 1\% degradation of the model performance, \sysname can drop the ASR of six attacks to less than 5\%.
However, the other method still struggles against the attacks, especially for Fine-Pruning.



\begin{table*}[]
\centering
\resizebox{0.9\textwidth}{!}{
\begin{tabular}{@{}c|cc|cc|cc|cc|cc|cc@{}}
\toprule
Attack & \multicolumn{2}{c|}{BadNet} & \multicolumn{2}{c|}{BA} & \multicolumn{2}{c|}{ETA} & \multicolumn{2}{c|}{IA} & \multicolumn{2}{c|}{SIG} & \multicolumn{2}{c}{TrojanNN} \\ \midrule
Method & ACC & ASR & ACC & ASR & ACC & ASR & ACC & ASR & ACC & ASR & ACC & ASR \\ \midrule
Before & 87.42 & 99.50 & 87.67 & 99.77 & 87.20 & 97.24 & 86.83 & 98.67 & 85.14 & 91.10 & 86.83 & 98.67 \\
Fine-Pruning + AM & 80.68 & 14.70 & 80.77 & 25.06 & 81.85 & 15.79 & 78.73 & 42.40 & 80.69 & 16.03 & 80.40 & 67.92 \\
Fine-Pruning + BS & 81.11 & 10.47 & 81.70 & 14.22 & 82.34 & 10.72 & 80.20 & 18.11 & 80.79 & 11.71 & 81.85 & 16.50 \\
Ours + AM & 81.72 & 12.35 & 82.03 & 17.28 & 81.97 & 12.78 & 81.51 & 30.55 & 81.88 & 15.17 & 83.09 & 55.32 \\
Ours + BS & \textbf{83.20} & \textbf{5.03} & \textbf{82.82} & \textbf{5.22} & \textbf{82.41} & \textbf{5.80} & \textbf{82.04} & \textbf{6.02} & \textbf{82.41} & \textbf{3.73} & \textbf{84.67} & \textbf{4.68} \\ \bottomrule
\end{tabular}
}
\caption{Comparison of backdoor defense with different purifying strategies and neuron importance evaluation metrics on CIFAR10.}
\label{table_importance_comp}
\end{table*}

\subsection{RQ2: A Closer Look at BS}
\label{bs_experiment}
As mentioned before, one of the key factors to make \sysname outperform other methods is the introduction of BS.
Here, comprehensive experiments are conducted to explain why BS can achieve such an improvement on backdoor defense from three perspectives: 1) statistical analysis; 2) attention analysis; 3) effect on neuron purifying.
In the experiments, a ResNet18 network is trained with CIFAR10.
During the training process, we backdoor the model according to the widely-used benchmark attack BadNet~\cite{Badnet} and ensure 100\% ASR. 
The poison label and poison ratio are set to 0 and 5\%. 
The trigger is a 3$\times$3 square in the upper left corner as suggested by~\cite{Badnet}.

\textbf{Statistical Analysis.}
We record the averaged BSs of all neurons by feeding a batch of clean and poison data, 500 for each, as shown in the left of Figure~\ref{fig_visualization}.
From the results, an interesting finding is that the BSs of neurons are almost opposite for clean and poison data.
In fact, such a phenomenon is precisely what we desire in bad neuron filtering.
This is because the phenomenon proves that BS well describes the principle of backdoor attacks: making some neurons to be strongly activated by triggers to change the original label output and rarely activated by clean data to avoid lowering the performance of victim model on normal inputs.
Also, we can use BS to accurately mark the neurons with low importance to normal predictions but high significance to trigger activation.
Referring to AM, shown in the right of Figure~\ref{fig_visualization}, it can be observed that there is no significant difference in AM between neurons.
Such a flaw makes AM unable to provide a clear division between benign and bad neurons, and leads the bad neurons marked by AM to be mixed in some benign neurons.

\textbf{Attention Analysis.}
Further, to validate the above statement, we visualize the attention of neurons on the original input image in Figure~\ref{fig_visualization}.
The results show that for both BS and AM, the attention of benign neurons is mostly focused on the main body of the input image, i.e.,around the airplane.
Then, the upper left corner, the place where triggers are added, draws the attention of bad neurons marked by BS.
Clearly, it is proved that with BS, \sysname accurately identifies which neurons contribute the most to backdoor attacks.
Conversely, the attention of the bad neurons marked by AM is not only focused on the upper left corner but also somewhere else.

\textbf{Effect on Neuron Purifying.}
Finally, we conduct experiments to show that the choice of neuron importance evaluation metric can lead to significant effect on neuron purifying.
From the results of Table~\ref{table_importance_comp}, given the same purifying method (Fine-Pruning or \sysname), the defense effect with BS outperforms the ones with AM on both ACC and ASR.
Especially, with AM, Fine-Pruning fails to defend the TrojanNN attack.
However, with BS, Fine-Pruning succeeds in defending TrojanNN due to the better ability of BS to identify bad neurons.
All the facts prove that BS is a good metric in backdoor defense.


\subsection{RQ3: Improvement of AR on Neuron Purifying}
To show the advantages of AR on fine-tuning based neuron purifying, we experimentally compare it with the other two commonly used regularization methods $L_1$ and $L_2$.
Figure~\ref{regularization} illustrates the performance of \sysname employed with three kinds of regularization mechanisms.


It is observed that the difference of these three kinds of regularization methods mainly stems from the changing trend of ASR.
In more details, the intrinsic property of $L_2$ makes it only ``strong'' on penalizing large parameters (more than 1, shown in Figure~\ref{diff_loss}), and unable to actually decrease parameters to 0..
However, for a backdoored network, the parameters of bad neurons are usually very small (close to 0).
Thus, $L_2$ fails to provide proper penalty on bad neurons and leads to the lowest ASR decreasing speed.
Then, for $L_1$, it can always reduce the ASR to a similar level to AR .
However, since the penalty of $L_1$ is too weak, its convergence speed is still lower than AR.
Finally, the required epochs of AR is just around 2, which is significantly lower than $L_1$ and $L_2$ regularization, which requires more than 15 epochs to erase the backdoor.
Thus, the adoption of AR in backdoor defense can greatly improve the performance of neuron purifying.

\begin{figure}[h]
\addtocounter{figure}{-1}
\centering
\subfigure{\begin{minipage}[t]{0.15\linewidth}\includegraphics[width=1.\linewidth]{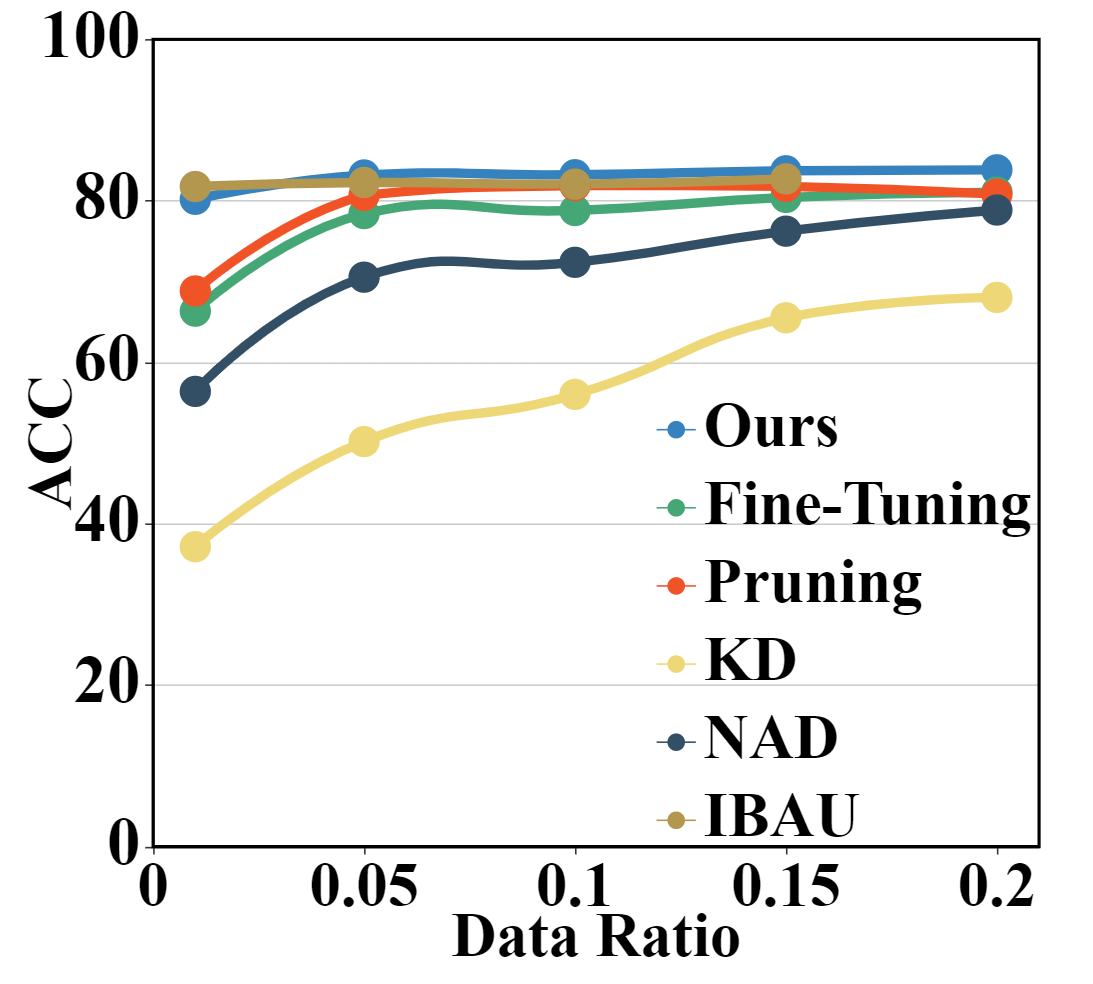}\end{minipage}}
\subfigure{\begin{minipage}[t]{0.15\linewidth}\includegraphics[width=1.\linewidth]{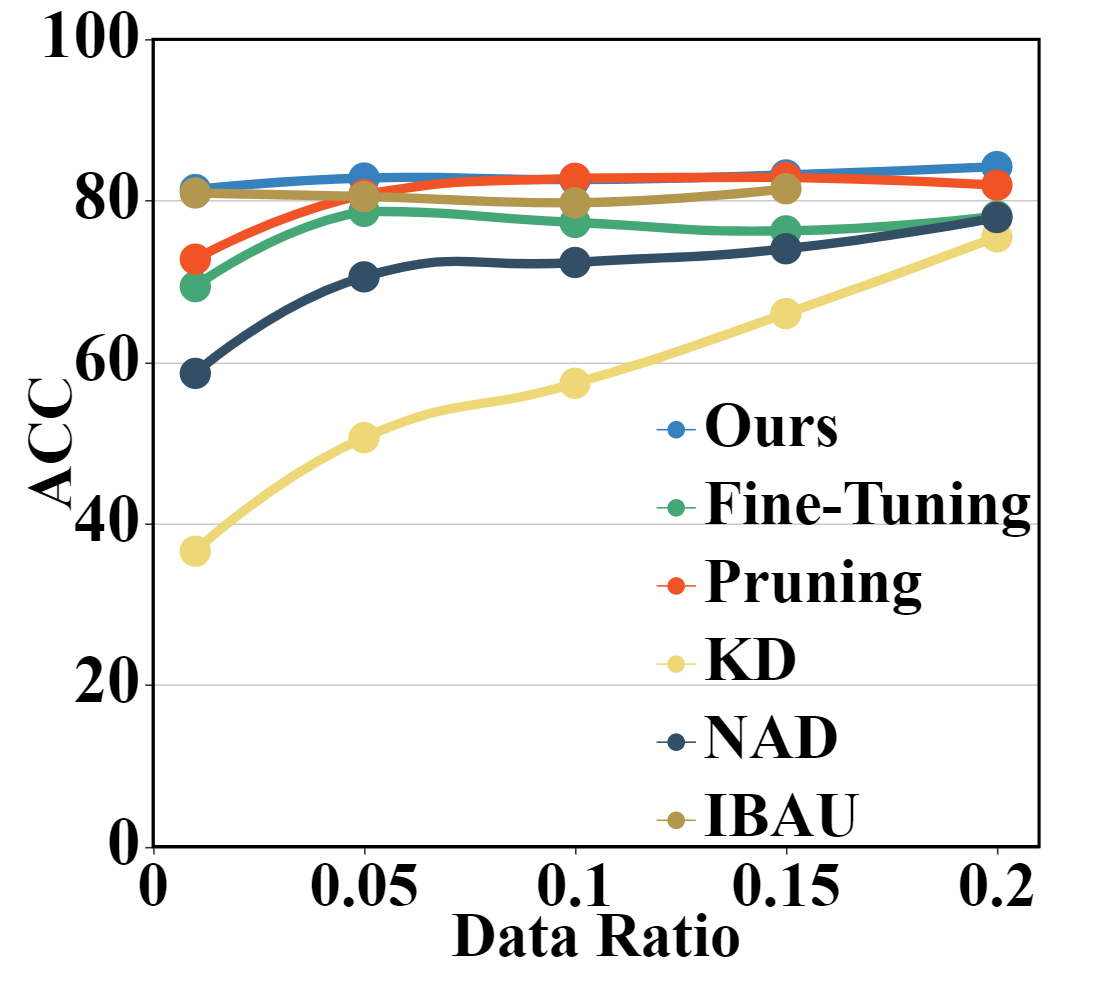}\end{minipage}}
\subfigure{\begin{minipage}[t]{0.15\linewidth}\includegraphics[width=1.\linewidth]{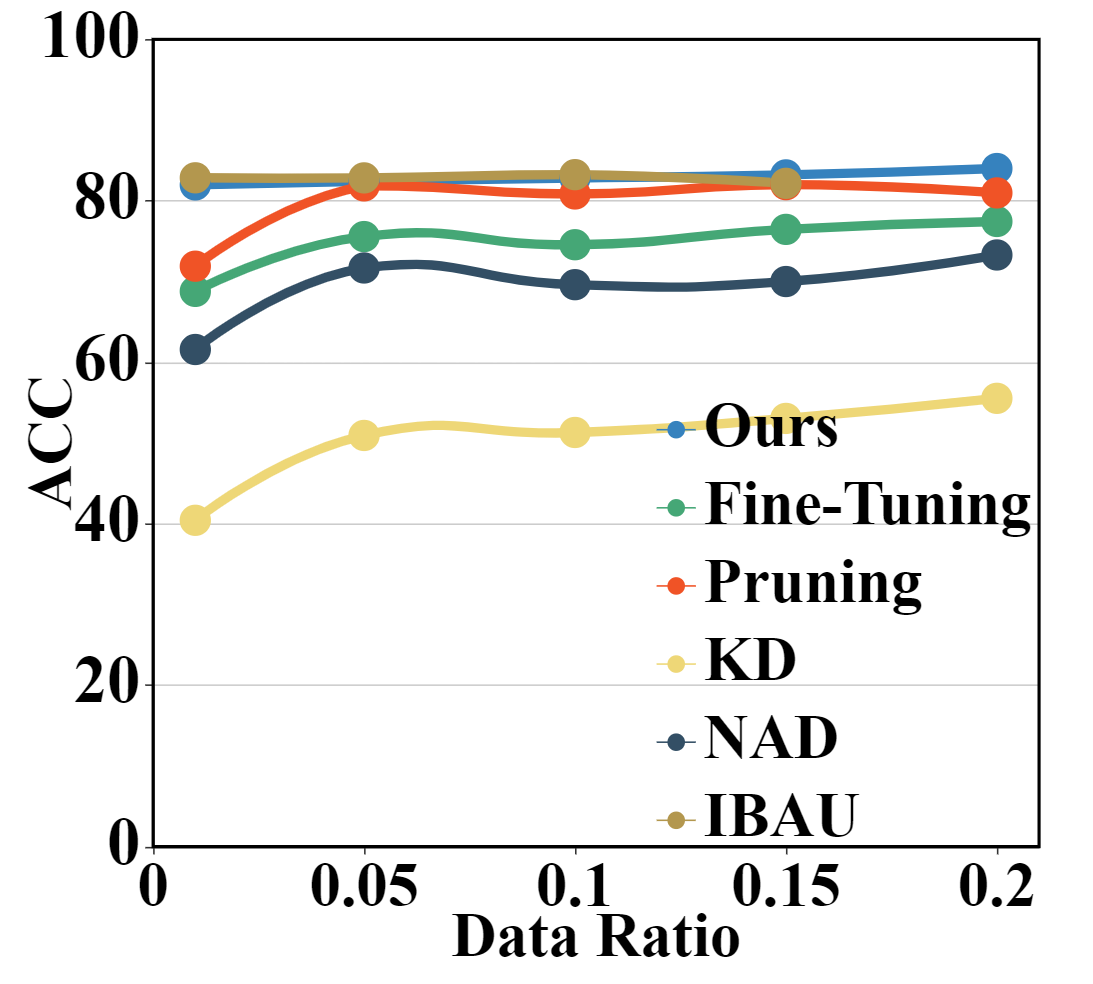}\end{minipage}}
\subfigure{\begin{minipage}[t]{0.15\linewidth}\includegraphics[width=1.\linewidth]{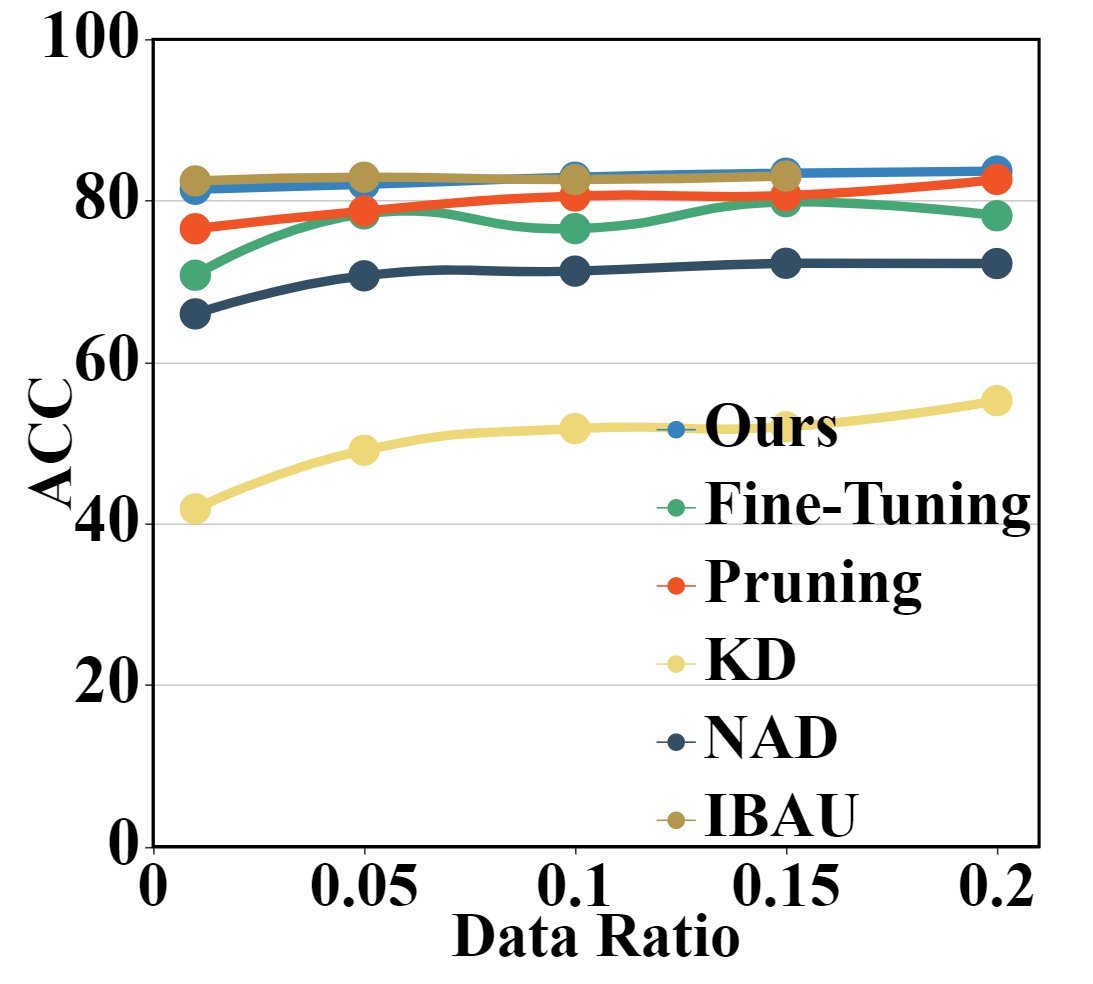}\end{minipage}}
\subfigure{\begin{minipage}[t]{0.15\linewidth}\includegraphics[width=1.\linewidth]{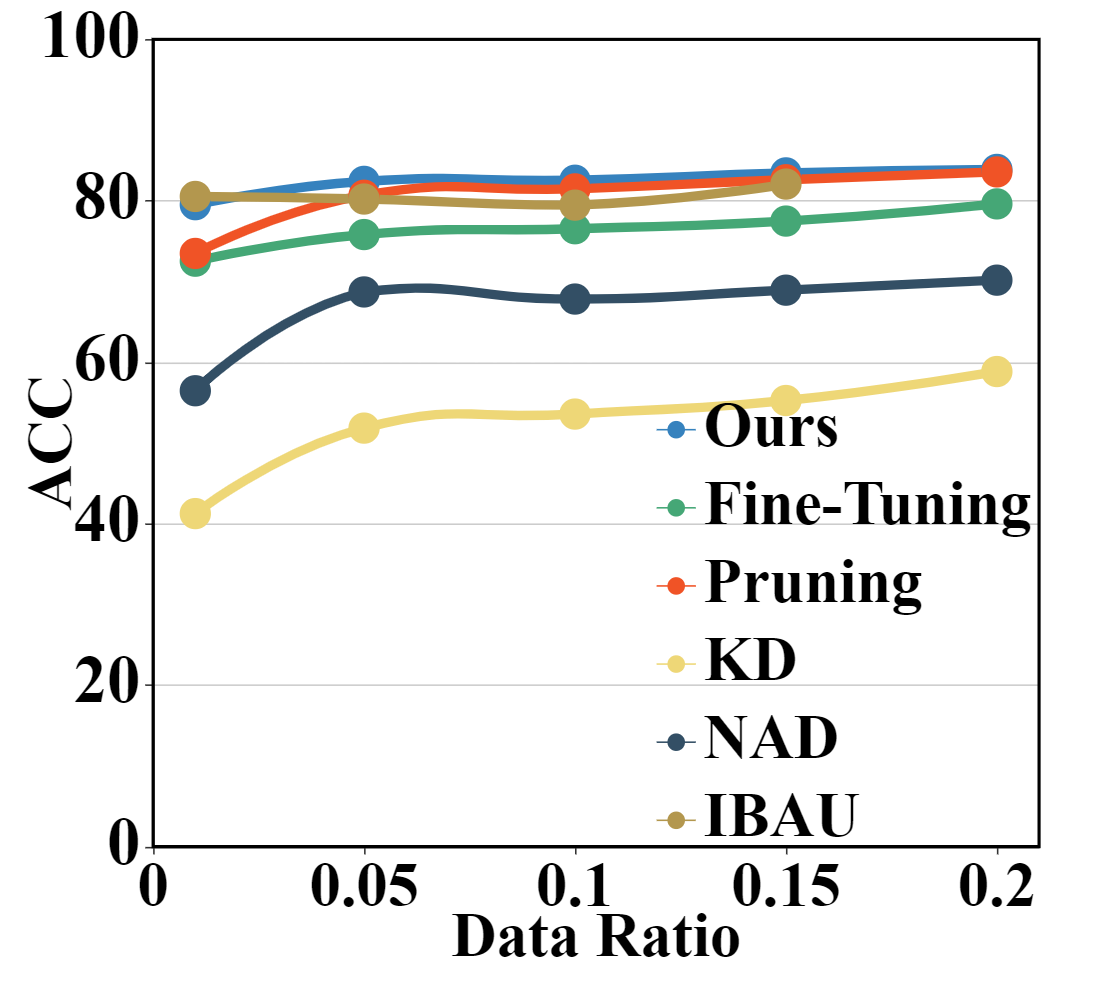}\end{minipage}}
\subfigure{\begin{minipage}[t]{0.15\linewidth}\includegraphics[width=1.\linewidth]{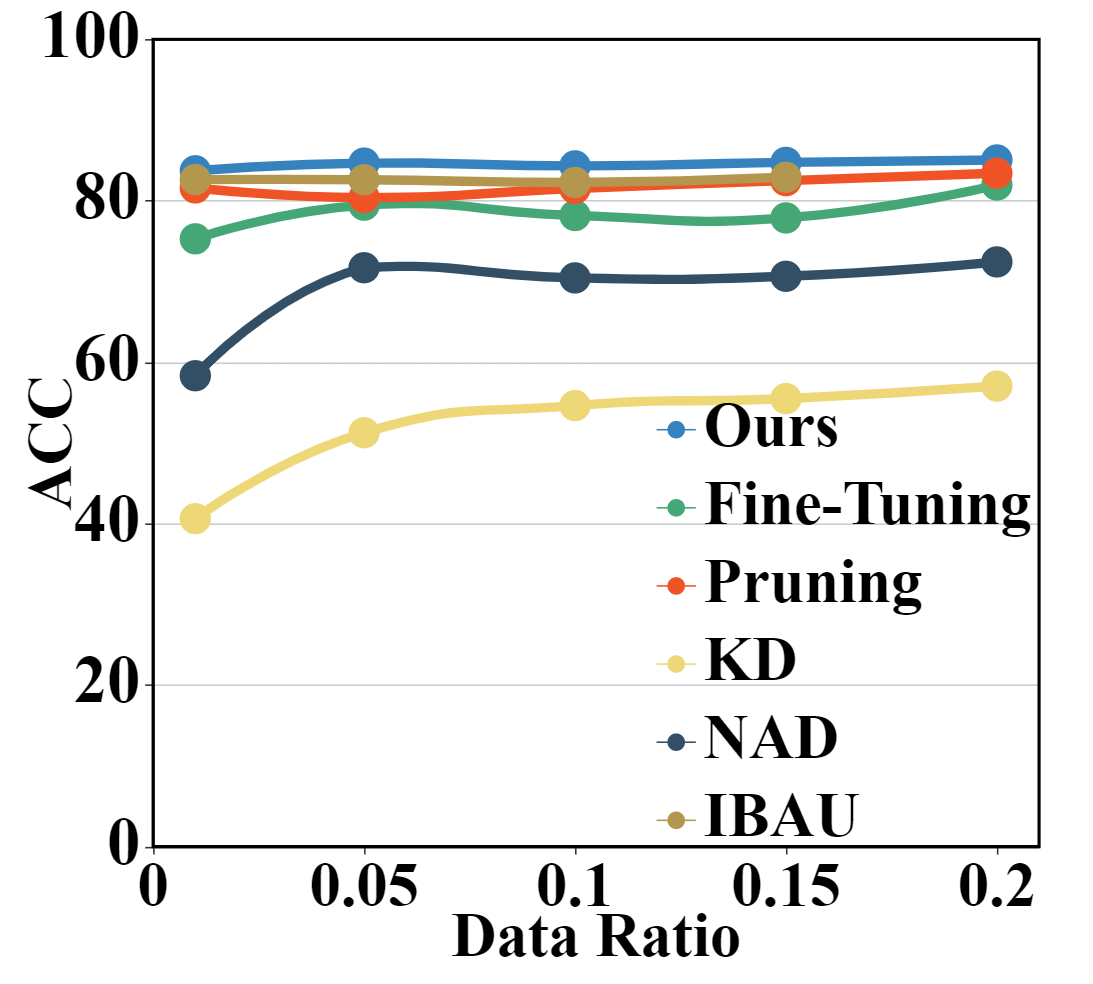}\end{minipage}}
\subfigure{\begin{minipage}[t]{0.15\linewidth}\includegraphics[width=1.\linewidth]{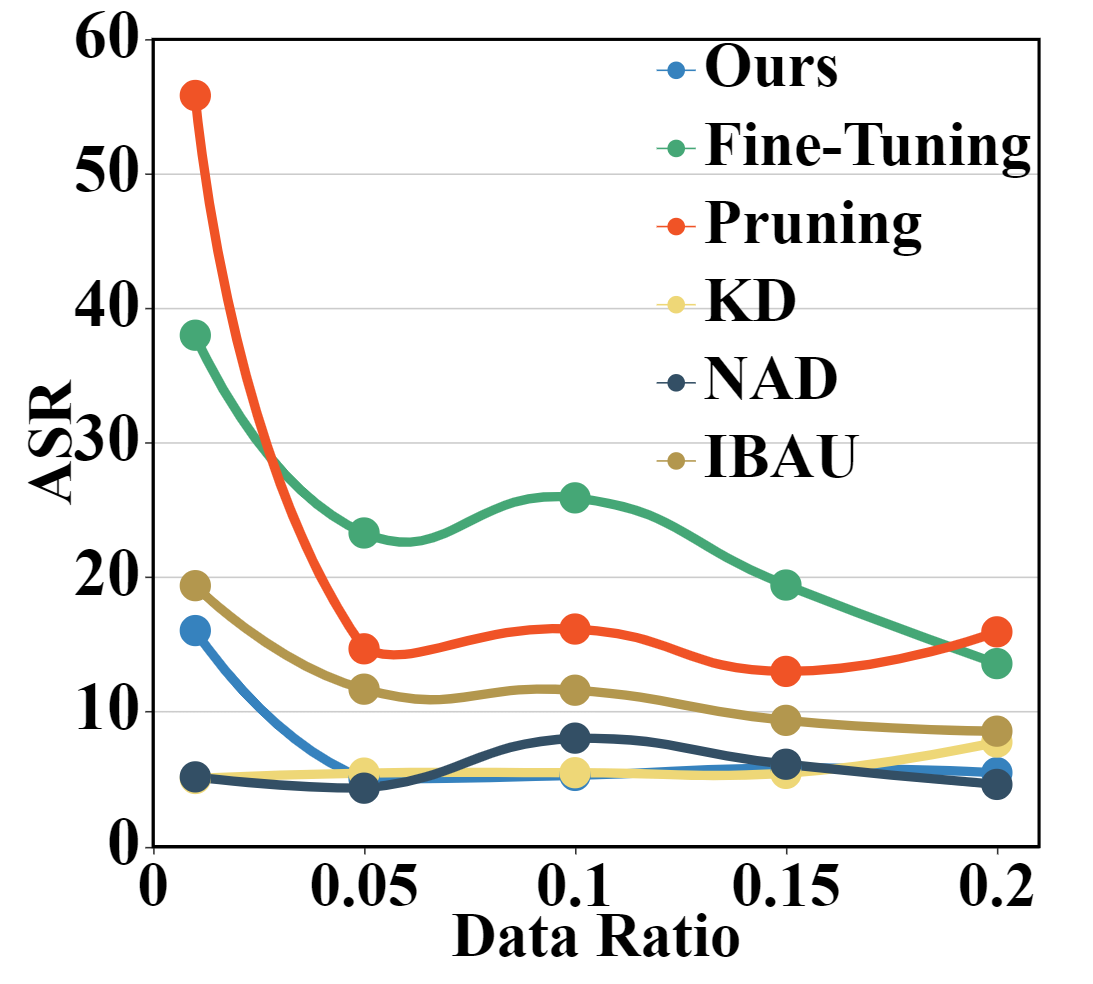}\end{minipage}}
\subfigure{\begin{minipage}[t]{0.15\linewidth}\includegraphics[width=1.\linewidth]{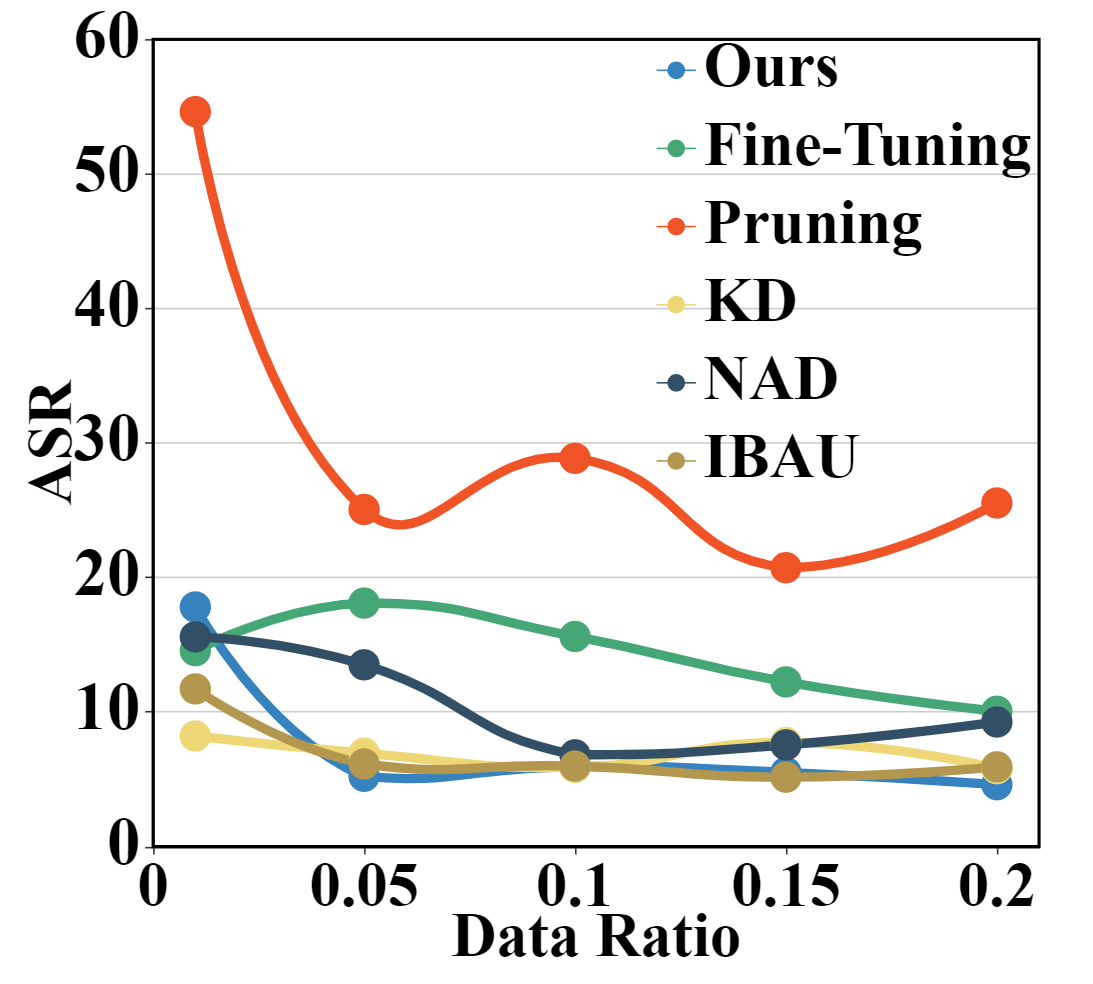}\end{minipage}}
\subfigure{\begin{minipage}[t]{0.15\linewidth}\includegraphics[width=1.\linewidth]{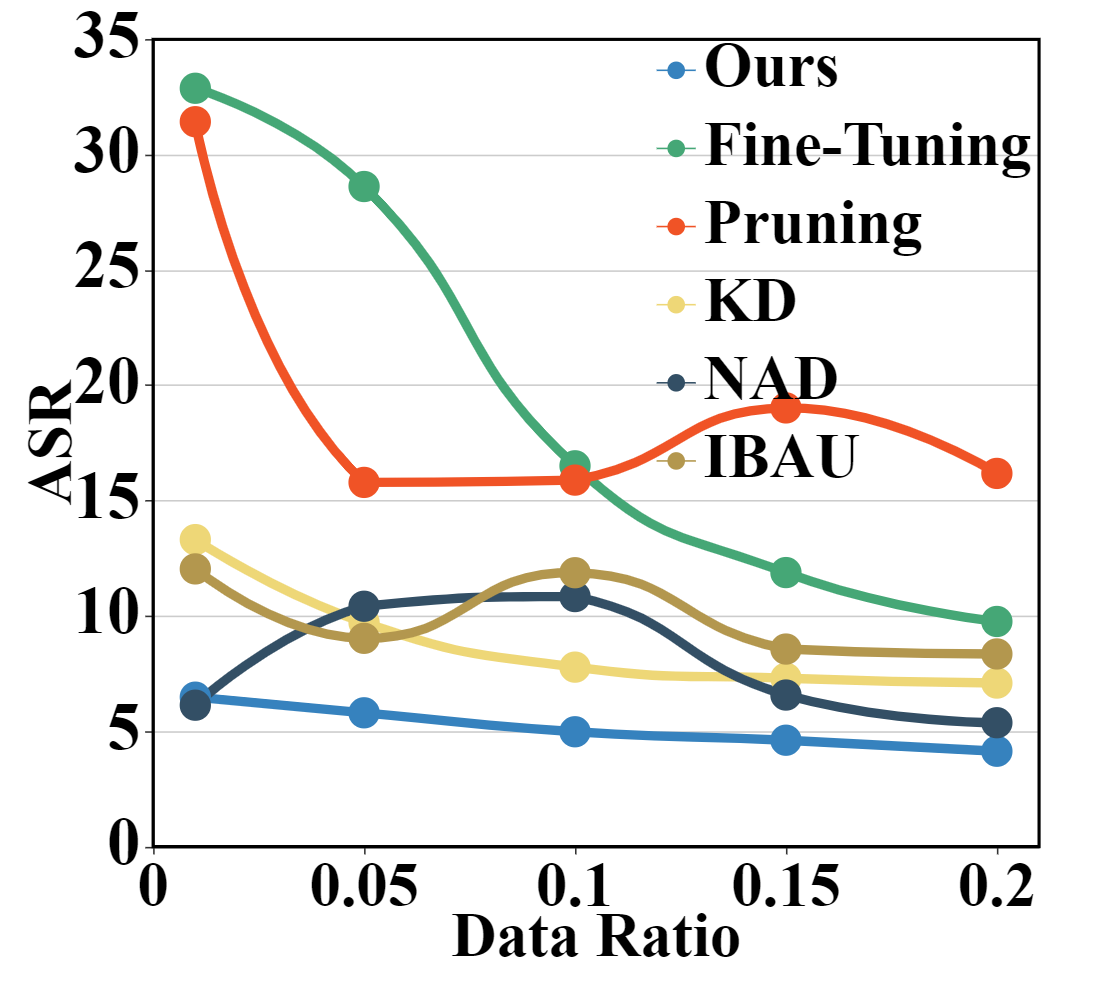}\end{minipage}}
\subfigure{\begin{minipage}[t]{0.15\linewidth}\includegraphics[width=1.\linewidth]{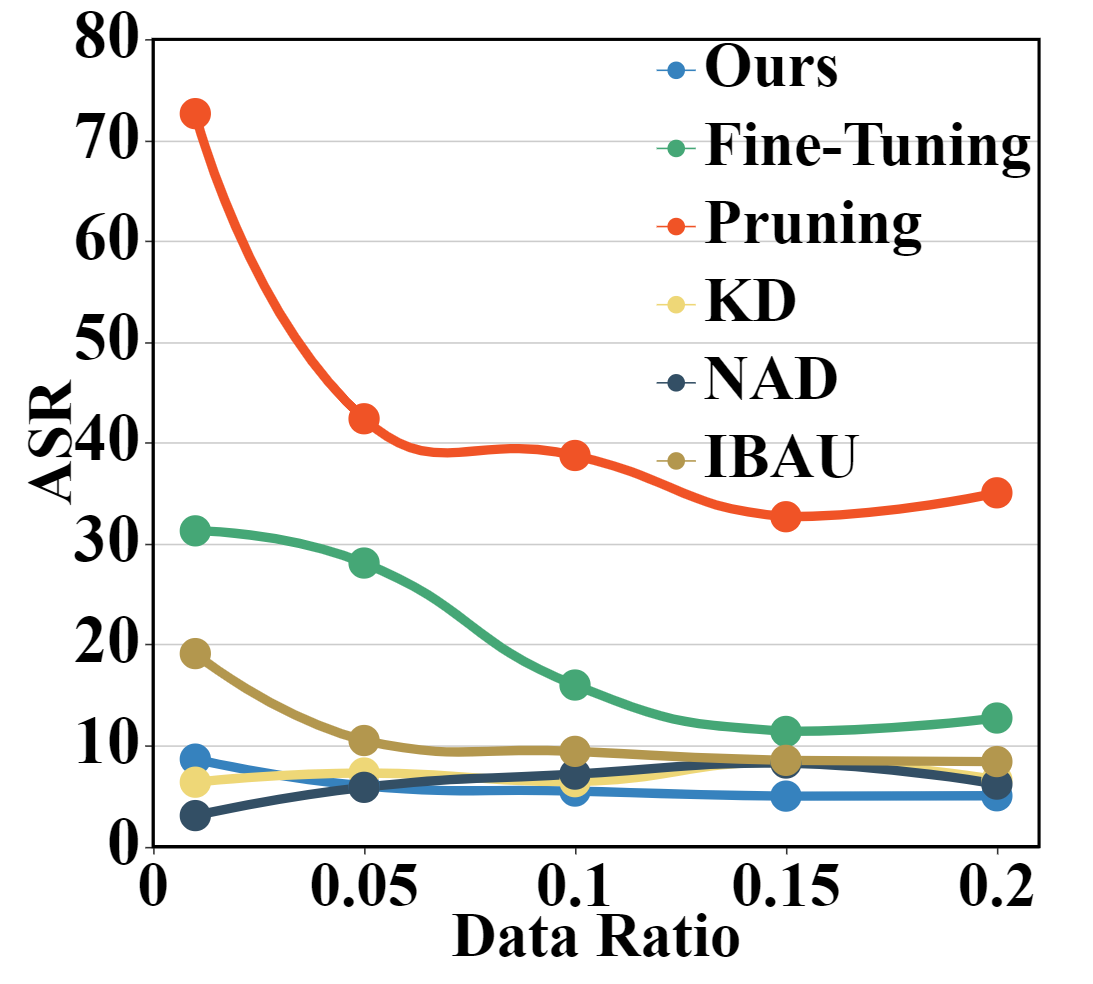}\end{minipage}}
\subfigure{\begin{minipage}[t]{0.15\linewidth}\includegraphics[width=1.\linewidth]{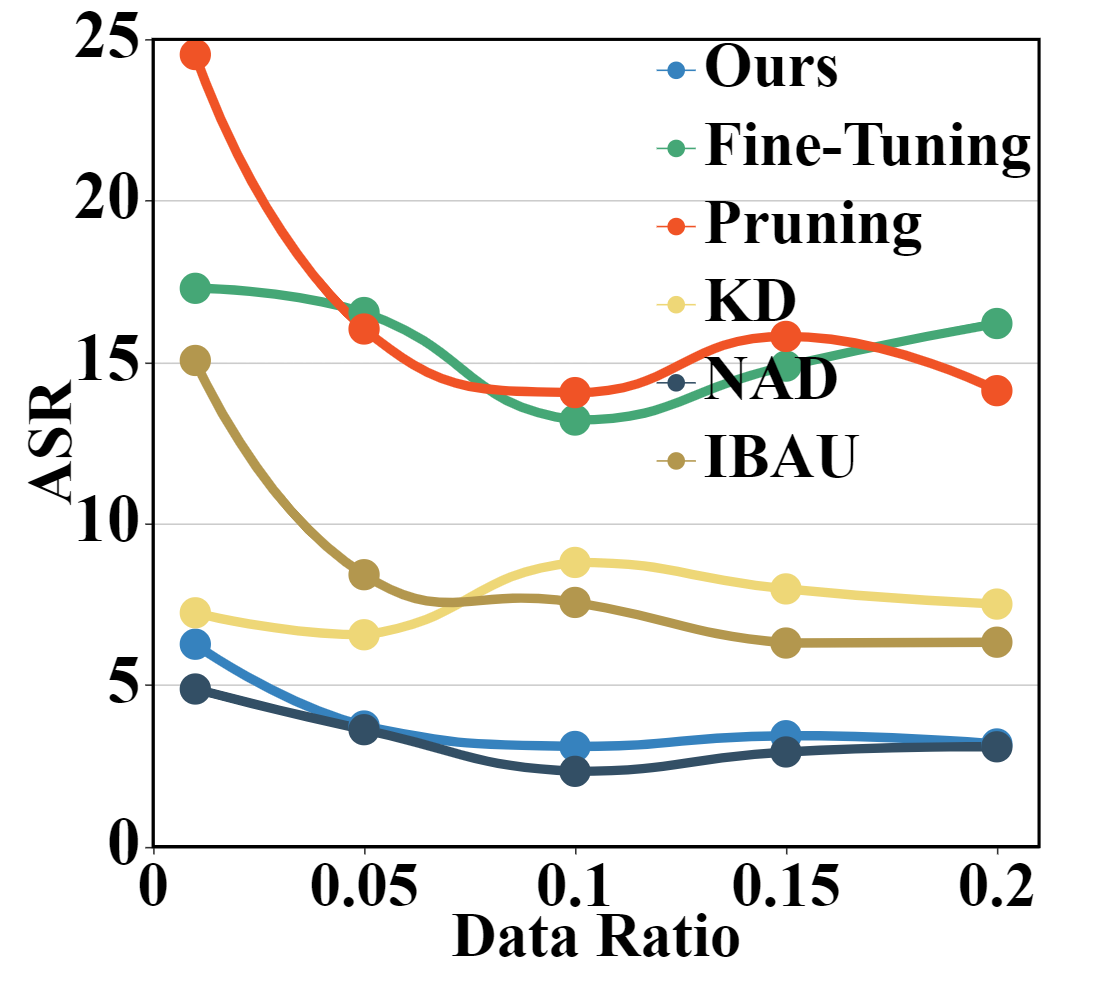}\end{minipage}}
\subfigure{\begin{minipage}[t]{0.15\linewidth}\includegraphics[width=1.\linewidth]{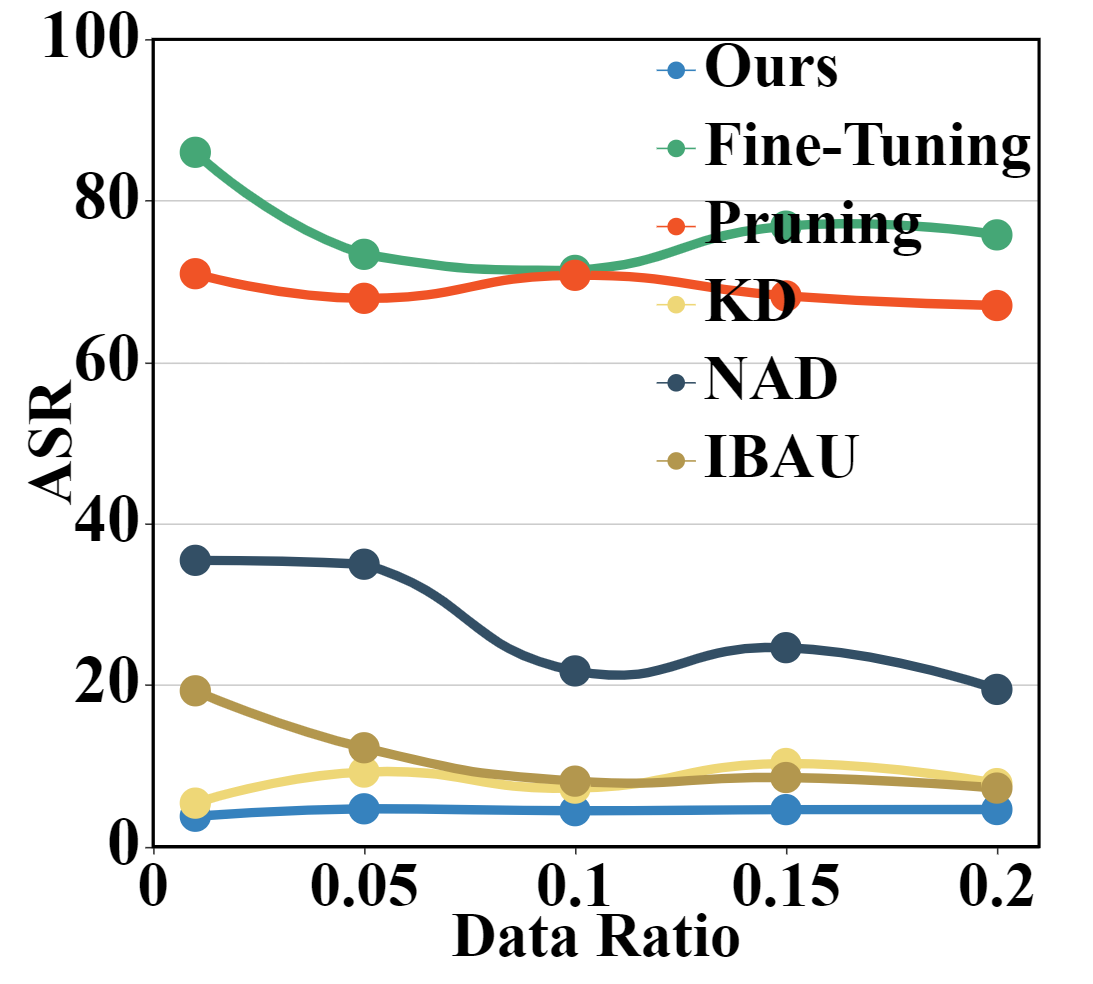}\end{minipage}}
\caption{The defense performance of five defense methods over different data ratio (0.01, 0.05, 0.1, 0.15, 0.2) in CIFAR-10. The images from left to right indicate against BadNet, BA, ETA, IA, SIG, and TrojanNN.}
\label{data_ratio}
\end{figure}


\begin{figure}[h]
\addtocounter{figure}{-1}
\centering
\subfigure{\begin{minipage}[t]{0.15\linewidth}\includegraphics[width=1.\linewidth]{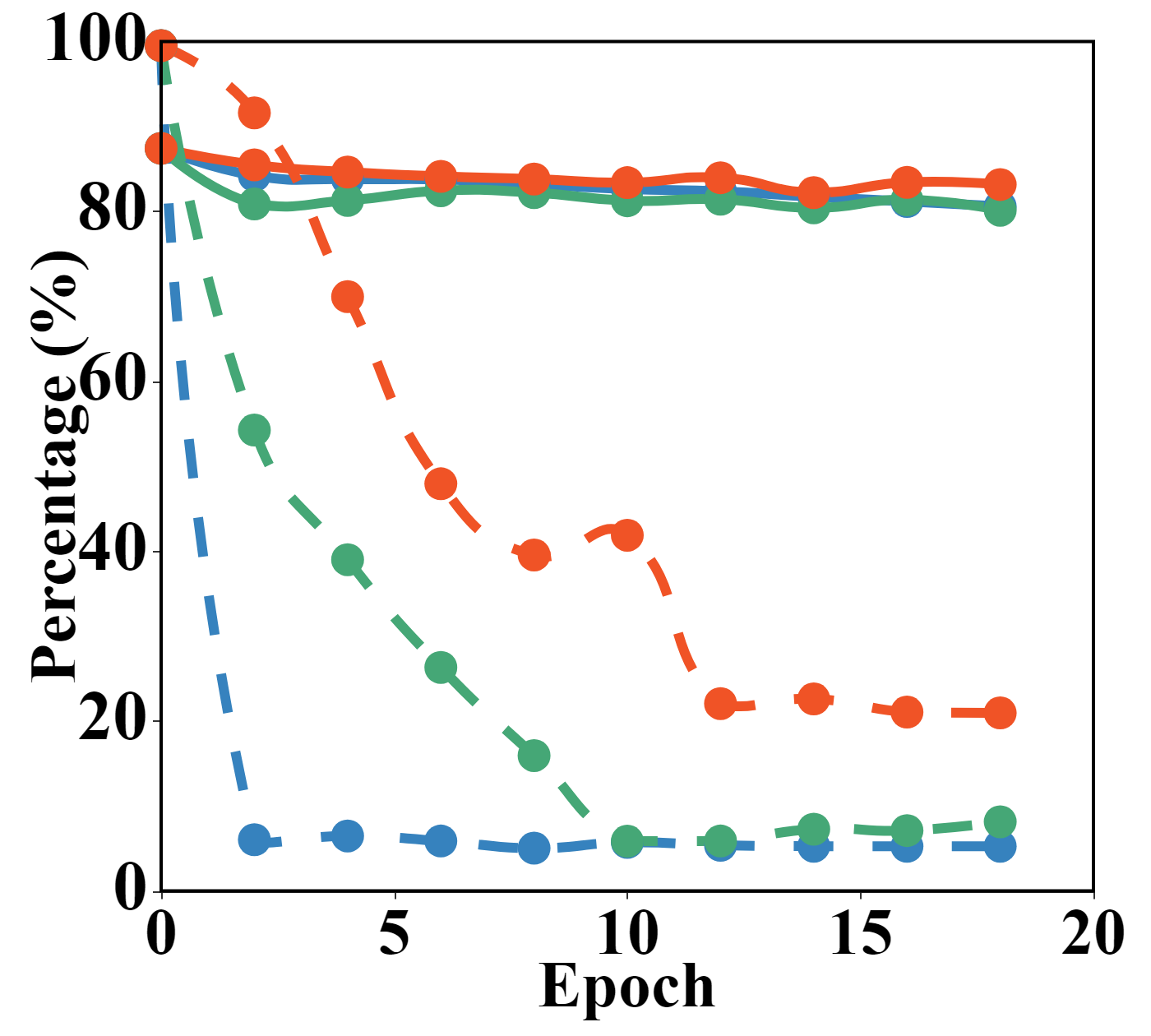}\end{minipage}}
\subfigure{\begin{minipage}[t]{0.15\linewidth}\includegraphics[width=1.\linewidth]{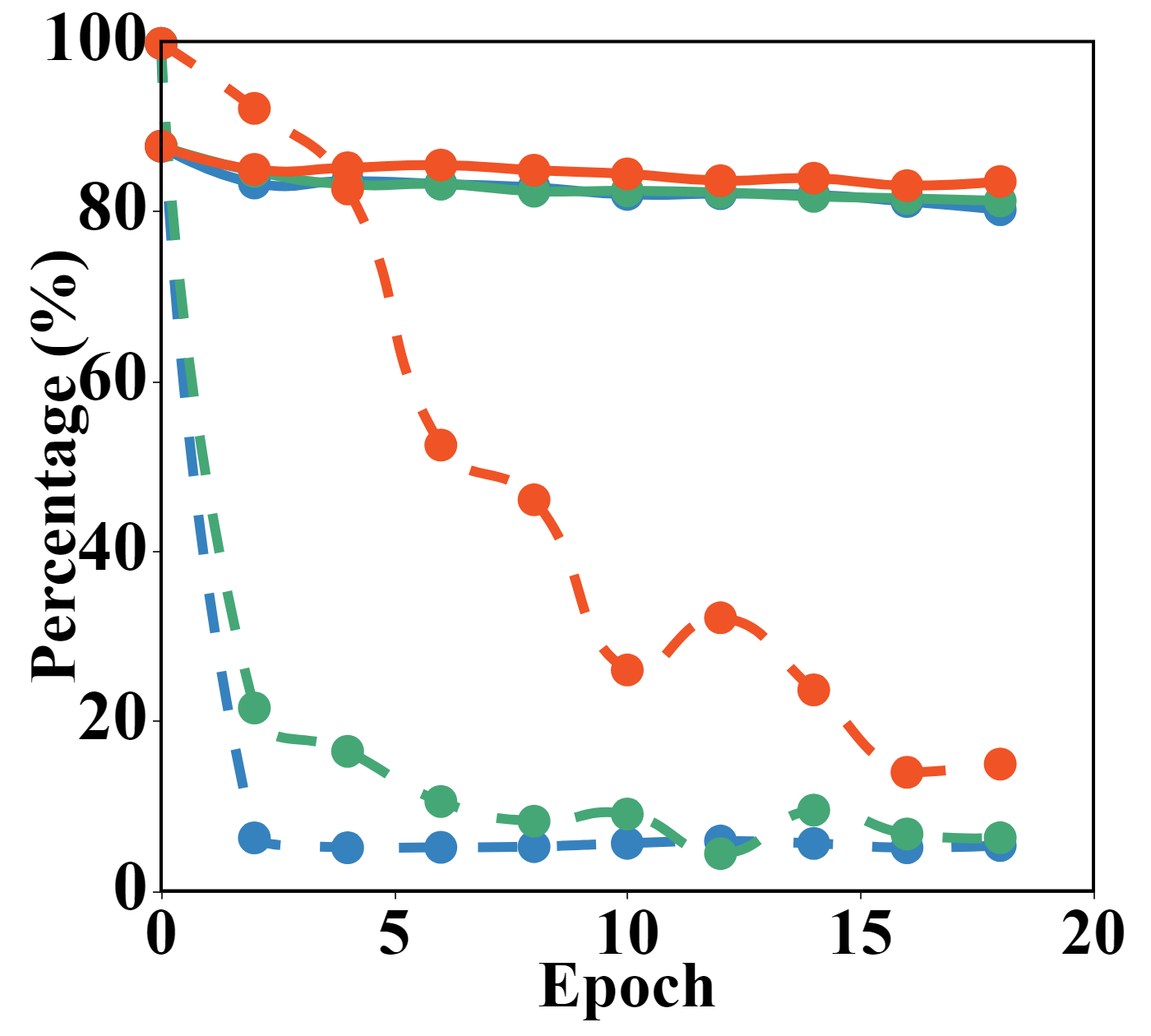}\end{minipage}}
\subfigure{\begin{minipage}[t]{0.15\linewidth}\includegraphics[width=1.\linewidth]{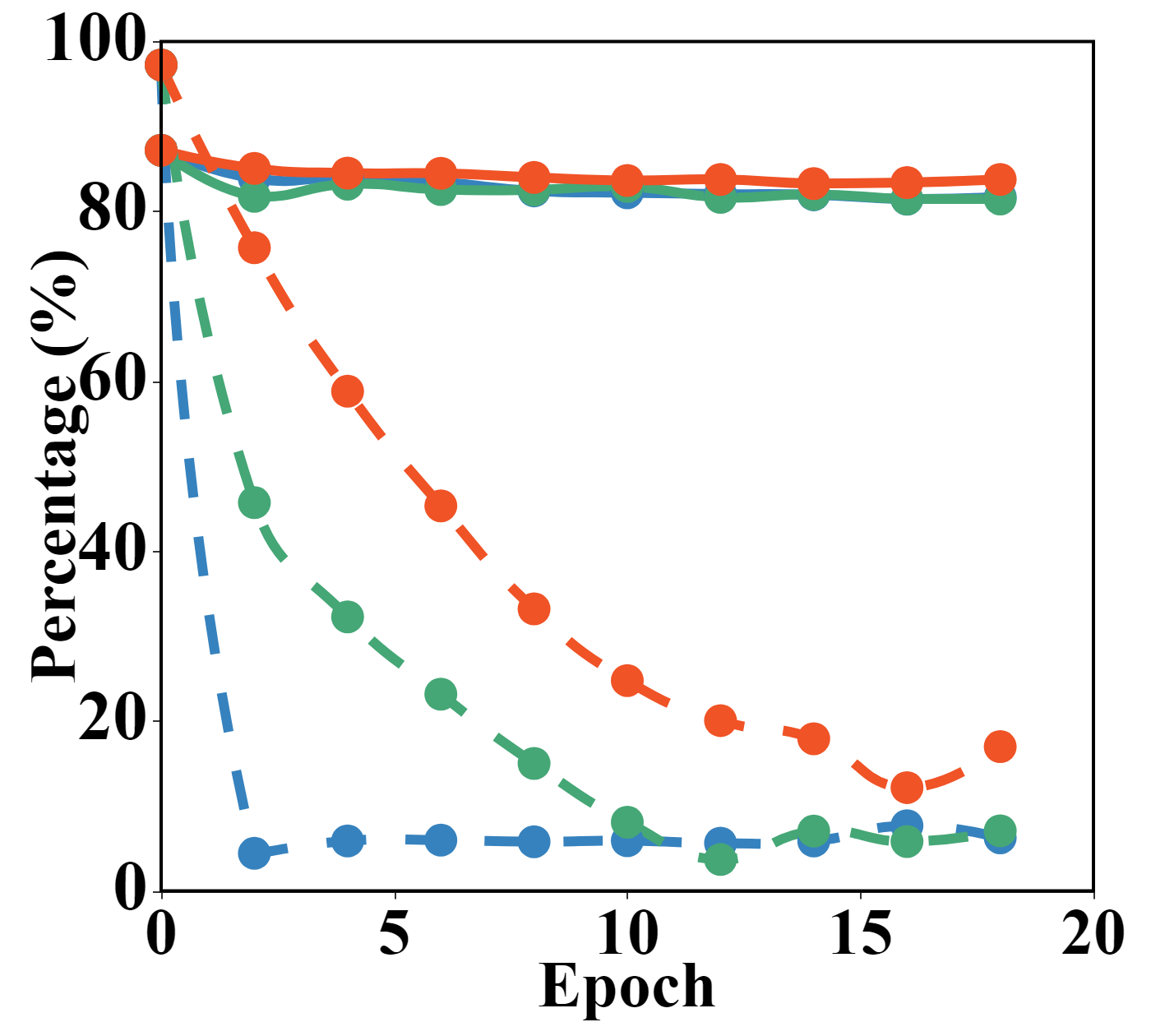}\end{minipage}}
\subfigure{\begin{minipage}[t]{0.15\linewidth}\includegraphics[width=1.\linewidth]{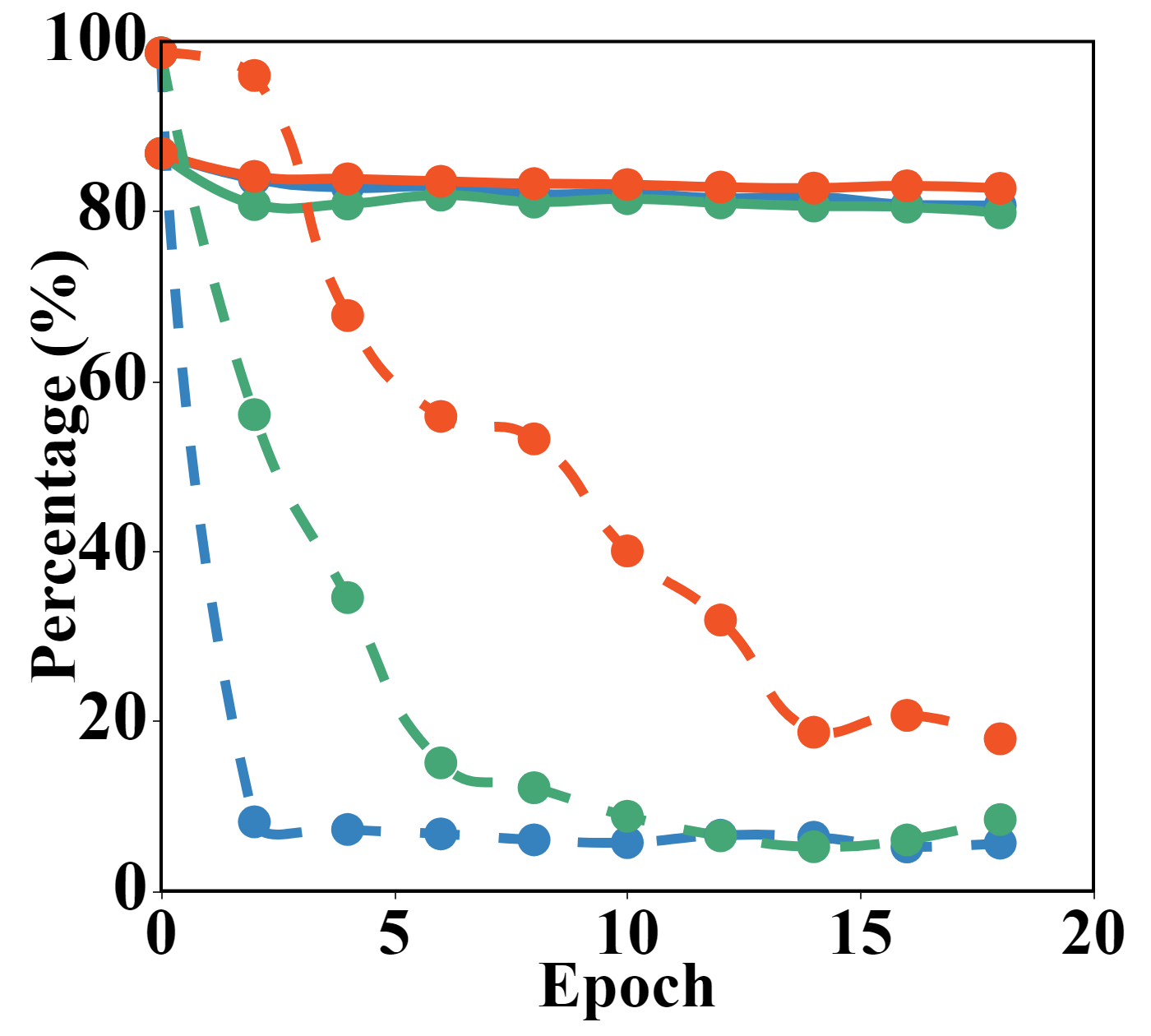}\end{minipage}}
\subfigure{\begin{minipage}[t]{0.15\linewidth}\includegraphics[width=1.\linewidth]{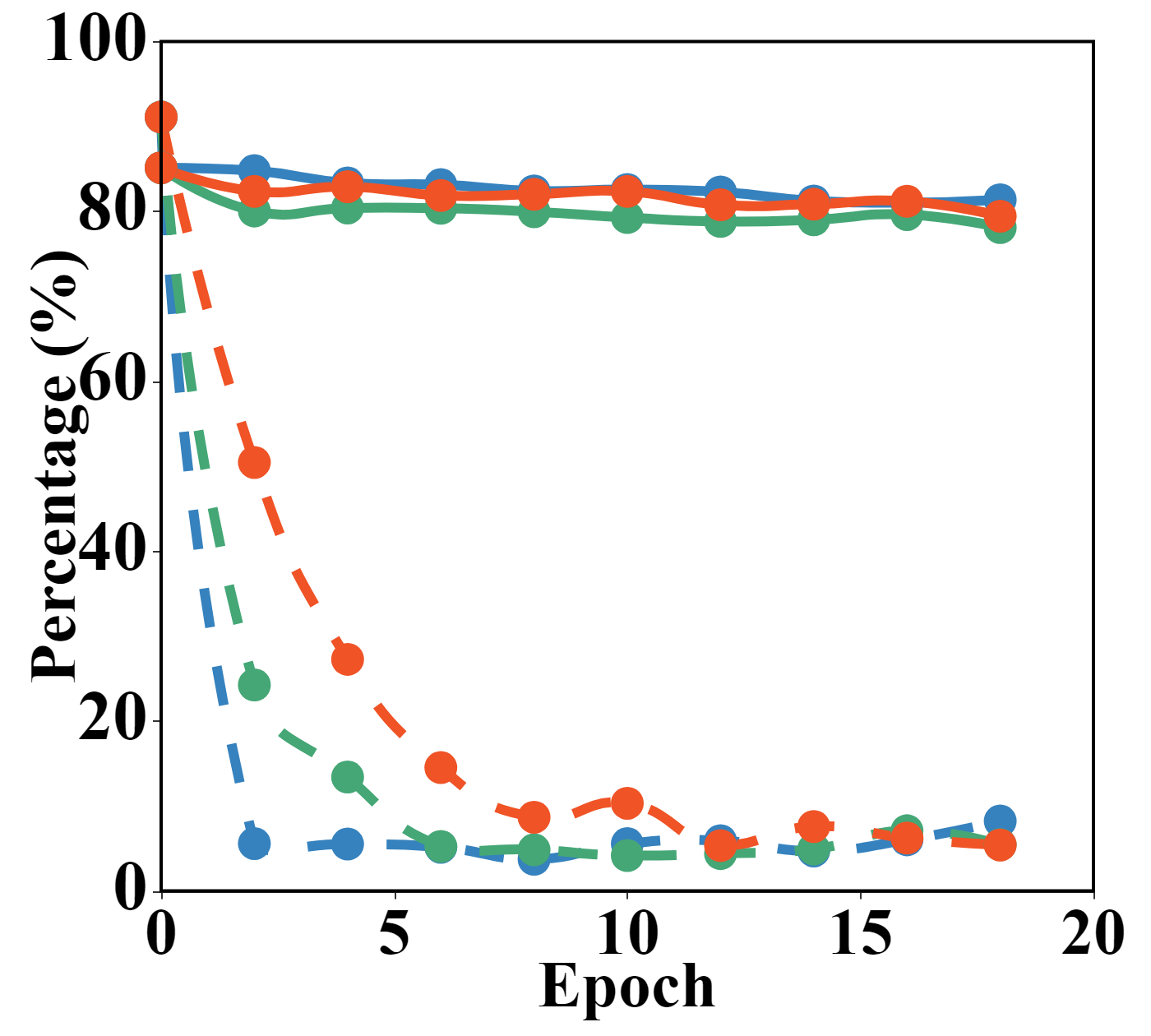}\end{minipage}}
\subfigure{\begin{minipage}[t]{0.15\linewidth}\includegraphics[width=1.\linewidth]{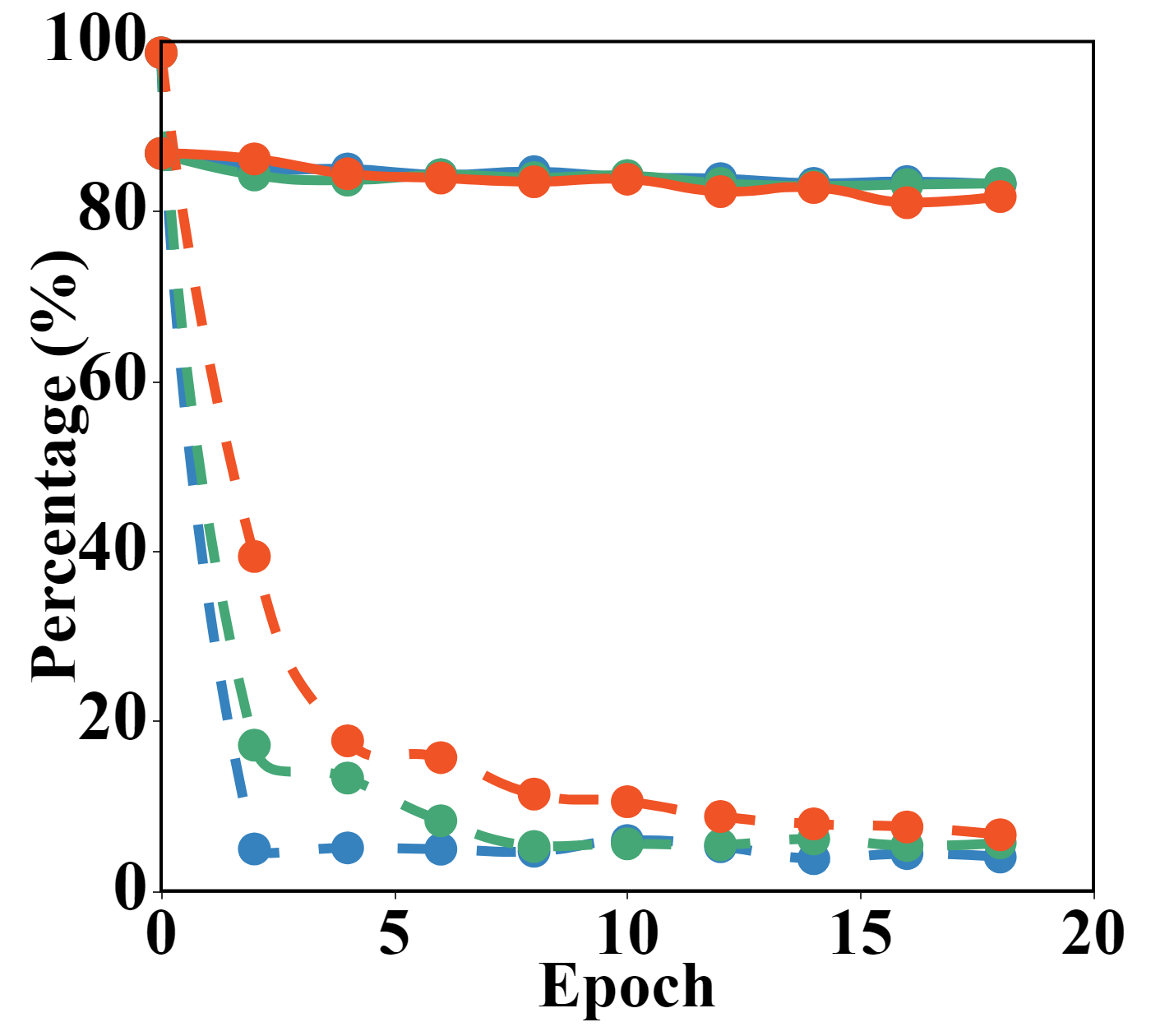}\end{minipage}}
\caption{The performance of \sysname with different regularization items ($L_1$, $L_2$, and $AR$) on CIFAR-10. The images from left to right indicate against BadNet, BA, ETA, IA, SIG, and TrojanNN.}
\label{regularization}
\end{figure}


\begin{table*}[!h]
\centering
\resizebox{0.75\textwidth}{!}{
\begin{tabular}{@{}c|cc|cc|cc|cc|cc|cc@{}}
\toprule
Attack & \multicolumn{2}{c|}{BadNet} & \multicolumn{2}{c|}{BA} & \multicolumn{2}{c|}{ETA} & \multicolumn{2}{c|}{IA} & \multicolumn{2}{c|}{SIG} & \multicolumn{2}{c}{TrojanNN} \\ \midrule
Alpha & ACC & ASR & ACC & ASR & ACC & ASR & ACC & ASR & ACC & ASR & ACC & ASR \\ \midrule
0.001 & \textbf{83.57} & 77.03 & \textbf{84.09} & 86.74 & 83.69 & 66.43 & \textbf{84.07} & 76.82 & \textbf{83.57} & 28.09 & 86.09 & 33.18 \\
0.005 & 83.55 & 21.01 & 84.14 & 26.20 & \textbf{84.41} & 16.54 & 83.44 & 25.97 & 82.64 & 6.69 & \textbf{86.22} & 5.57 \\
0.01 & 83.20 & \textbf{5.03} & 82.82 & 5.22 & 82.41 & 5.80 & 82.04 & 6.02 & 82.41 & 3.73 & 84.67 & 4.68 \\
0.05 & 82.76 & 5.48 & 82.89 & 5.42 & 82.10 & 5.48 & 81.97 & 5.69 & 81.78 & 4.02 & 84.25 & 4.86 \\
0.1 & 82.20 & 5.30 & 82.18 & \textbf{5.07} & 81.55 & \textbf{3.96} & 81.33 & \textbf{4.72} & 81.20 & \textbf{3.44} & 84.39 & \textbf{4.39} \\ \bottomrule
\end{tabular}
}
\caption{The performance of \sysname with different $\alpha$ against six backdoor attacks in CIFAR-10.}
\label{alpha}
\end{table*}

\subsection{RQ4: Is purifying output layers enough to defend backdoor attacks}\label{sub_rq4}
In this subsection, we mainly answer the question of whether only purifying the last few layers is enough for erasing the backdoor from the victim model.
Table \ref{purifying_different_layer} reports the performance of \sysname as the target purifying layers are changed.
From the results, adding feature extraction layer (i.e., convolution layers) into backdoor purifying shows little improvement on defense effect, only less 1\% more decrease on ASR.
In contrast, the performance of the purified model degrades more significantly when more convolution layers are introduced.
As aforementioned, such a phenomenon occurs because backdoor attacks usually pay less attention on the feature extraction layers than the output layer to achieve better attack effectiveness.
Therefore, forcibly purifying the feature extraction layers contributes little to the backdoor defense but can easily lead to the over-pruning of the victim model and lower the model performance.

\begin{table*}[]
\centering
\resizebox{0.95\textwidth}{!}{
\begin{tabular}{@{}c|cc|cc|cc|cc|cc|cc@{}}
\toprule
Attack         & \multicolumn{2}{c|}{BadNet}    & \multicolumn{2}{c|}{BA} & \multicolumn{2}{c|}{ETA} & \multicolumn{2}{c|}{IA} & \multicolumn{2}{c|}{SIG}       & \multicolumn{2}{c}{TrojanNN} \\ \midrule
Purified Layer & ACC            & ASR           & ACC             & ASR            & ACC                  & ASR                 & ACC               & ASR              & ACC            & ASR           & ACC              & ASR             \\ \midrule
FC only        & \textbf{83.20} & 5.03          & \textbf{82.82}  & 5.22           & \textbf{82.41}       & 5.80                & \textbf{82.04}    & 6.02             & \textbf{82.41} & 3.73          & \textbf{84.67}   & 4.68            \\
FC + 1 Conv    & 82.19          & 5.15          & 81.29           & 5.93           & 81.37                & 5.77                & 80.46             & 5.44             & 81.26          & 3.98          & 82.89            & 4.24            \\
FC + 2 Conv    & 80.29          & 4.79          & 79.31           & 5.24           & 79.42                & 5.38                & 77.66             & 5.69             & 79.82          & 4.05          & 81.00            & \textbf{4.01}   \\
FC + 3 Conv    & 77.03          & 4.83          & 77.84           & \textbf{4.34}  & 77.65                & 4.60                & 76.71             & 5.62             & 77.96          & 3.45          & 79.72            & 5.36            \\
FC + 4 Conv    & 74.49          & \textbf{4.51} & 74.86           & 4.35           & 71.56                & \textbf{4.04}       & 73.33             & \textbf{4.96}    & 74.56          & \textbf{3.22} & 72.73            & 4.67            \\ \bottomrule
\end{tabular}
}
\caption{The performance of \sysname when purifying different layers. Here $n$ Conv indicates purifying the last n convolution layers.}
\label{purifying_different_layer}
\end{table*}

\subsection{Other Cases}
\label{ablation_study}
To better understand \sysname, we perform extensive ablation studies with varying experimental settings, namely data ratio and punishment coefficient $\alpha$ .

\textbf{Impact of Clean Data.}
As discussed in the prior works~\cite{fine_pruning,fine_tuning,KD,NAD}, the amount of clean data owned by the defender is a crucial factor that affects the effect of backdoor defense significantly.
Here, for the convenience of statistic and comparison, we use the ratio of clean data set to training set as the indicator to denote the amount of clean data owned by the defender.
Figure \ref{data_ratio} demonstrates the performance of five defense methods against six attacks over different data ratios.
Strikingly, with the strict condition of 1\% data ratio, \sysname can still maintain almost 90\% ASR drops and less than 3.5\% accuracy drops.
In contrast, most of other methods fail to achieve effective backdoor defense (less than 40\% ASR drops on some attacks) or suffer from significant ACC drops (more than 20\%).

\textbf{Impact of $\alpha$.}
As discussed in Section~\ref{sub_method}, $\alpha$ is a vital factor to determine the degree of \sysname to punish the bad neurons.
Table~\ref{alpha} summarizes the experimental results of \sysname with coarse-tuning of $\alpha$.
The results show that larger $\alpha$ makes \sysname focus more on backdoor purifying and achieve lower ASR but is with the expense of slightly decreased ACC.
Correspondingly, lower $\alpha$ allows the target model to maintain higher ACC but results in a higher ASR.
Therefore, a trade-off exists between ASR and ACC.
However, the trade between ASR and ACC are not exchanged at equal values.
It can be found that in most cases, only less than 1\% ACC drop can leads to the drop of ASR for about 50\% to 60\%.
As a result, in practice, the defender is suggested to choose a litter higher magnitude of $\alpha$ to ensure the effectiveness of \sysname.

\textbf{Impact of Purifying Strategy.}
Furthermore, in Table~\ref{table_importance_comp}, we also show the influence of purifying strategy choice on backdoor defense.
As expected, despite of maintaining the performance of the purified model in an acceptable range, Fine-Pruning (fine-tuning with $L_1$ + pruning) still suffers from poor capacity of backdoor erasing.
However, if Fine-Pruning is replaced with our purifying strategy, the results can be improved significantly.
For instance, referring to the TrojanNN attack, substituting Fine-Pruning with our purifying strategy can make the ACC rise from 81.85\% to 84.67\% while the ASR drops from 16.50\% to 4.68\%, even with AM as the importance evaluation metric.


\section{Conclusion}
\label{conclusion}
In this paper, we proposed a backdoor defense approach \sysname that could combine the advantages of both model pruning and fine-tuning.
In more details, \sysname mainly improved current backdoor defense method from two aspects.
The first was to propose a new metric called benign salience to evaluate the importance of neurons in a network and filter the bad ones.
The second was to design a new regularization item for backdoor purifying, which could accelerate the neuron purifying process significantly.
Finally, we conducted extensive experiments to show that the \sysname achieved ideal performance on both erasing the backdoor and maintaining the model performance after being equipped with such two techniques.
The codes would be opensourced after this paper is publicly available.


\bibliographystyle{iclr2021_conference}
\bibliography{iclr2021_conference}

\end{document}